# Understanding and Modeling Perceived Cognitive and Physical Strain Dynamics for Planning-Oriented Human–Robot Collaboration in Prefabricated Construction

Yifan Wang[1], Bo Xiao[2*], Shane T. Mueller[3]

1. Ph.D. Candidate, Department of Civil, Environmental, and Geospatial Engineering, Michigan Technological University, Houghton, 49931 MI, USA. Email: wyifan@mtu.edu
2. Assistant Professor, Department of Civil, Environmental, and Geospatial Engineering, Michigan Technological University, Houghton, 49931 MI, USA. Email: boxiao@mtu.edu
3. Professor, Department of Psychology and Human Factors, Michigan Technological University, Houghton, 49931 MI, USA. Email: shanem@mtu.edu

**Abstract:** Human–robot collaboration (HRC) in prefabricated construction requires planning approaches that consider not only productivity but also time-dependent worker states during repeated work and rest. Existing planning models often rely on simplified assumptions about fatigue, workload, or recovery, with limited domain-specific empirical evidence on how perceived strain evolves. This study develops an empirically grounded, planning-oriented approach to characterize perceived strain accumulation and recovery in prefabricated construction HRC. A controlled repeated work–rest experiment assessed perceived cognitive and physical strain using the Rating Scale for Mental Effort and Borg's Rating of Perceived Exertion. Linear and exponential functional forms were evaluated, followed by mixed-effects modeling to examine collaborative conditions, session effects, and inter-individual variability. Results indicate that cognitive strain accumulation is best represented by a linear mixed-effects model, whereas rest-phase recovery follows nonlinear decay. The resulting planning-oriented models may inform future human-state-aware task allocation and scheduling research.

## 1. INTRODUCTION

Prefabricated construction has become an important approach for improving productivity, quality, and schedule reliability by shifting substantial construction activities from variable field environments to more standardized factory-based settings [1]. In these settings, prefabricated components or modules are commonly manufactured through repeated and sequential workflows before being transported for on-site assembly. These workflows often involve material handling, framing, fastening, and inspection, where production efficiency depends on well-coordinated task sequencing, resource allocation, and schedule control [2]. To further improve productivity, flexibility, and safety, collaborative robots and human–robot teams have been increasingly explored in prefabricated construction by combining robotic precision and repeatability with human adaptability [3–5].

Human–robot collaboration (HRC) task allocation and scheduling are therefore central to maintaining production flow, reducing delays, and balancing productivity with worker well-being in prefabricated construction systems [6,7]. Conventional planning and scheduling models often implicitly treat human execution as time-invariant, with limited consideration of how worker states and performance fluctuate during extended work and rest periods [8–10]. In practice, workers may experience increasing cognitive effort during inspection or monitoring tasks and increasing physical effort during repetitive manual operations. When sustained over extended work periods, these effort demands may contribute to fatigue and subsequent performance decline. Insufficient representation of these human-state dynamics may lead to task allocation decisions that overlook worker-state changes, potentially increasing safety risks and affecting the long-term reliability of HRC systems.

Prior research on construction robotics and HRC has examined multiple human-centered topics related to task allocation and scheduling, including safety-aware interaction, adaptive collaboration, human-state sensing, and training [6,11–13]. These studies demonstrate the growing recognition that human factors are important for safe and effective HRC. Nevertheless, existing HRC planning models often rely on simplified assumptions, such as fixed fatigue-related constraints, constant productivity rates, or preset accumulation and recovery parameters [14–16]. Although such assumptions may be useful for model formulation, their transferability can be limited when functional forms or parameters are not empirically calibrated for the specific task context, worker population, and work–rest structure.

Cognitive resource theories, including Multiple Resource Theory, have long recognized that sustained task demands can impose competing resource requirements, contributing to progressive strain and fatigue-related outcomes [17–19]. In this study, perceived strain refers to workers' self-reported cognitive and physical effort in response to task demands. Fatigue is treated as a longer-term outcome associated with sustained effort and resource expenditure, whereas recovery refers to the reduction of perceived strain during rest. In prefabricated construction HRC, however, limited empirical evidence is available on how perceived cognitive and physical strain accumulates during work and recovers during rest. Although many HRC experiments measure performance, workload, fatigue or physiological responses, fewer are designed to generate repeated work–rest observations suitable for identifying planning-oriented accumulation and recovery patterns [20–22]. As a result, planning models still lack empirically grounded representations of how perceived worker state changes over time during HRC operation.

Treating perceived cognitive and physical strain as time-dependent worker states, this study develops a planning-oriented modeling approach to empirically characterize human-state dynamics for prefabricated construction HRC. First, the study provides an experimentally grounded protocol that enables repeated observation of perceived strain accumulation and recovery within a representative HRC workflow. Second, it empirically evaluates linear and exponential functional forms and mixed-effects structures to describe how perceived cognitive and physical strain evolves over time under human-only and collaborative conditions. Third, we discuss how these models may inform future human-centered task allocation and scheduling by representing worker state as a time-dependent process that can empirically modeled under controlled task conditions. By empirically examining perceived strain dynamics under repeated work–rest conditions, this work bridges human factors measurement and construction robotics planning and provides a foundation for future human-aware HRC scheduling research.

## 2. LITERATURE REVIEW

### 2.1 Human–robot collaboration in prefabricated construction: applications, human roles, and scheduling needs

Prefabricated construction generally refers to a construction method in which building components are produced in factory-based or off-site environments and then transported to the construction site for assembly [1]. Compared with conventional field construction, prefabricated construction involves more standardized workflows, repeated task patterns, controlled workspaces, and scalable production [2]. These characteristics have made prefabrication an important approach for improving productivity and reducing lead time. For example, in high-rise public housing projects in Hong Kong, contractors have adopted a six-day cycle since the 1990s for the on-site assembly of a typical floor, usually consisting of 20–30 prefabricated units [23]. At the same time, factory-based prefabrication creates strong requirements for automation and production coordination.

Construction-specific HRC research has increasingly explored how robots can assist with prefabricated or semi-structured construction tasks. A recent review of HRC in modular construction manufacturing identified 78 relevant publications, indicating growing research interest in applying HRC to modular and prefabricated production [3]. Liu et al. proposed a robotic installation system for assembling prefabricated components, including precast concrete beams, slabs, walls and columns [24]. Yang et al. developed an assembly sequence planning method for factory-based robotic timber wall prefabrication, including stud placement and fastening [25]. While these studies demonstrate the technical feasibility of applying robots to prefabricated construction tasks, they provide relatively limited discussion of human roles and capabilities within collaborative workflows.

Human roles have also been increasingly emphasized in recent human-centered HRC studies in prefabricated construction. Even in prefabrication factories, construction materials and components are not always as uniform as products in conventional manufacturing. Material variability, workflow transitions, and project-specific design may still require human judgment and adjustment [13,26]. Yang et al. highlighted that HRC in timber prefabrication can combine robot precision and human flexibility, offering opportunities for productivity and flexibility beyond traditional automation [27]. In this division of work, robots are well suited for repetitive, positioning-intensive, or physically demanding subtasks, whereas human workers often remain responsible for monitoring, inspection, quality control, task transitions, and handling unexpected variations [28]. These studies demonstrate that safe and effective HRC depends not only on robot capability, but also on human flexibility and adaptation.

Task allocation and scheduling are therefore especially important in HRC systems in prefabricated construction because human and robot actions must be coordinated across sequential workflows. Prior studies have shown that appropriate task allocation and scheduling can improve productivity [29], enhance safety [12], and balance human factors [30]. Poor allocation or timing may increase worker or robot idle time, underuse robotic resources, overload workers, or reduce workflow reliability. For instance, Tsai et al. developed and implemented an iterative HRC framework in a timber wall panel fabrication process, improving the overall process by 34% in the first iteration and 13% in the second [29]. Tehrani and Alwisy proposed a Human-in-the-Lead Construction Robotics (HiLCR) framework for wood assembly that optimizes task allocation by prioritizing human leadership [6]. Together, these studies demonstrate that task allocation and scheduling are important for coordinating human–robot workflows in prefabricated construction.

## 2.2 Human-state representation models in planning-oriented HRC research

Recent research on HRC planning has increasingly emphasized the need to move beyond purely task-centric planning and scheduling toward human-centric decision-making [9,28,31]. Traditional HRC planning models usually address how task sequences are assigned and scheduled within human–robot teams to improve efficiency under resource constraints. In these models, human workers are often represented as homogeneous and time-invariant resources, typically characterized by fixed processing times, static availability, or constant productivity rates [16,32,33].

For example, a disassembly sequence planning algorithm for industrial HRC [8] and a deep multi-agent reinforcement learning–based scheduling approach for HRC assembly [7] both focus on optimizing task assignment efficiency while implicitly assuming time-invariant human execution characteristics. Although such assumptions are computationally tractable, they provided limited representation of temporal variability in workers' performance caused by adaptation, learning, fatigue, and human–robot interaction.

To address this limitation, human-state-aware planning models has begun to incorporate fatigue, physical capability, mental workload, execution competency, and multimodal human-state estimation into HRC planning and scheduling. Exponential fatigue formulations are commonly adopted to represent human state evolution, such as the Learning-Forgetting-Fatigue-Recovery (LFFR) model [34]. For instance, an LFFR-based multi-objective optimization framework was developed to balance cycle time and personalized physical fatigue in HRC assembly [14]. In this study, inter-worker differences were characterized through muscle strength measurements, but fatigue accumulation and recovery rates were still treated as temporally fixed during repeated operations. Other studies have expanded human state representation beyond physical fatigue. A proactive HRC framework examined mutual cognitive, predictable and self-organizing perspectives [31], while Yao et al. proposed a task reallocation optimization framework that integrates multimodal data analysis to update fatigue estimates dynamically [15]. More recently, Zhao et al. incorporated physical exertion intensity, mental exertion intensity, and execution competency into human-centric assembly planning [9]. Together, these studies demonstrate the feasibility of incorporating diverse human-state variables into HRC planning models.

Industrial HRC and manufacturing research therefore provides useful supporting context for this study [15,31,35]. These studies demonstrate that human-state variables can be mathematically represented and incorporated into optimization models. However, industrial manufacturing contexts differ from prefabricated construction in task scale, workspace variability, material handling requirements, and the continuing need for human judgment in partially automated workflows. Therefore, industrial HRC models are valuable references, but their assumptions and parameters require careful evaluation before being applied to HRC systems in prefabricated construction. Table 1 summarizes representative human-state variables and limitations in recent planning-oriented HRC research.

**Table 1.** Comparison of prior studies on human-state representation in planning-oriented HRC research

| Study | Human-state variable | Model / approach | Key limitation for this study's objective |
|---|---|---|---|
| Chand and Lu [14] | Physical fatigue and recovery | LFFR-based multi-objective scheduling optimization | Considers personalized fatigue, but fatigue/recovery rates are model-based rather than calibrated from repeated work–rest observations. |
| Yao et al. [15] | Dynamic mental fatigue from multimodal data | Fatigue evaluation network and dynamic task reallocation | Updates fatigue estimates for reallocation, but focuses on fatigue detection/updating rather than functional-form identification. |
| Zhao et al. [9] | Physical exertion, mental exertion, and body posture | Human-centric assembly planning with multi-objective optimization | Integrates multiple human factors into planning, but does not empirically model work–rest human-state dynamics. |
| Baratta et al. [35] | Operator fatigue and productivity | Simulation-based HRC task allocation | Evaluates fatigue and productivity at scenario level; |

| | | | limited empirical calibration of worker-state dynamics |
|---|---|---|---|
| Present study | Perceived cognitive and physical strain | Linear/exponential functional form comparison and mixed-effects modeling | Provides empirically grounded, planning-oriented strain models; future field validation and optimization integration are needed. |

The comparison shows that prior planning-oriented HRC studies have incorporated human-related factors into task allocation and scheduling, but many still rely on simplified model structures, assumed parameters, or outcome-derived updates. Such assumptions may limit transferability to prefabricated construction HRC, where human-state dynamics can depend on the specific workflow, collaboration mode, worker population, and work–rest structure. Therefore, the key limitation is not the use of exponential fatigue or recovery models themselves, but the lack of empirical calibration and model identification for the target task context. In this study, a functional form refers to the mathematical shape used to describe how perceived strain changes over time. For example, linear accumulation assumes a constant rate of increase during work, whereas exponential recovery assumes rapid early reduction followed by slower later recovery during rest.

**2.3 Theoretical framing of workload, perceived strain, fatigue, and recovery**

Multiple Resource Theory and resource depletion/recovery perspectives provide the theoretical basis for treating perceived cognitive and physical strain as related but partially distinct worker-state processes. Multiple Resource Theory suggests that task demands compete for limited cognitive resources depending on processing modality, task modality, and response requirements [36]. In ergonomics and human factors research, sustained physical work is similarly constrained by strength, endurance, metabolic limits, and recovery capacity [37,38]. These perspectives indicate that cognitive and physical demands may draw on different resource dimensions and therefore may show different accumulation and recovery patterns during HRC tasks.

In this study, workload refers to task-imposed cognitive or physical demand. Inspection and monitoring tasks mainly impose cognitive workload, whereas material handling and repetitive manual operations impose physical workload. Effort refers to the cognitive or physical resources expended by a worker in response to task demands [39]. Accordingly, this study uses perceived strain to denote workers' self-reported cognitive or physical effort in response to those demands. Resource depletion refers to the latent reduction in available cognitive or physical capacity caused by sustained resource use. Within this framework, fatigue is viewed as a longer-term outcome of sustained effort and resource expenditure, whereas recovery refers to the reduction of perceived strain during rest or reduced-demand periods [40,41].

Figure 1 summarizes this conceptual relationship. Task workload generates perceived cognitive or physical strain, which reflects workers' reported effort in response to task demands. Sustained strain may contribute to resource depletion and fatigue-related outcomes, while rest supports partial recovery by reducing perceived strain. The figure also distinguishes cognitive and physical pathways to reflect the possibility that they may follow different accumulation and recovery patterns.

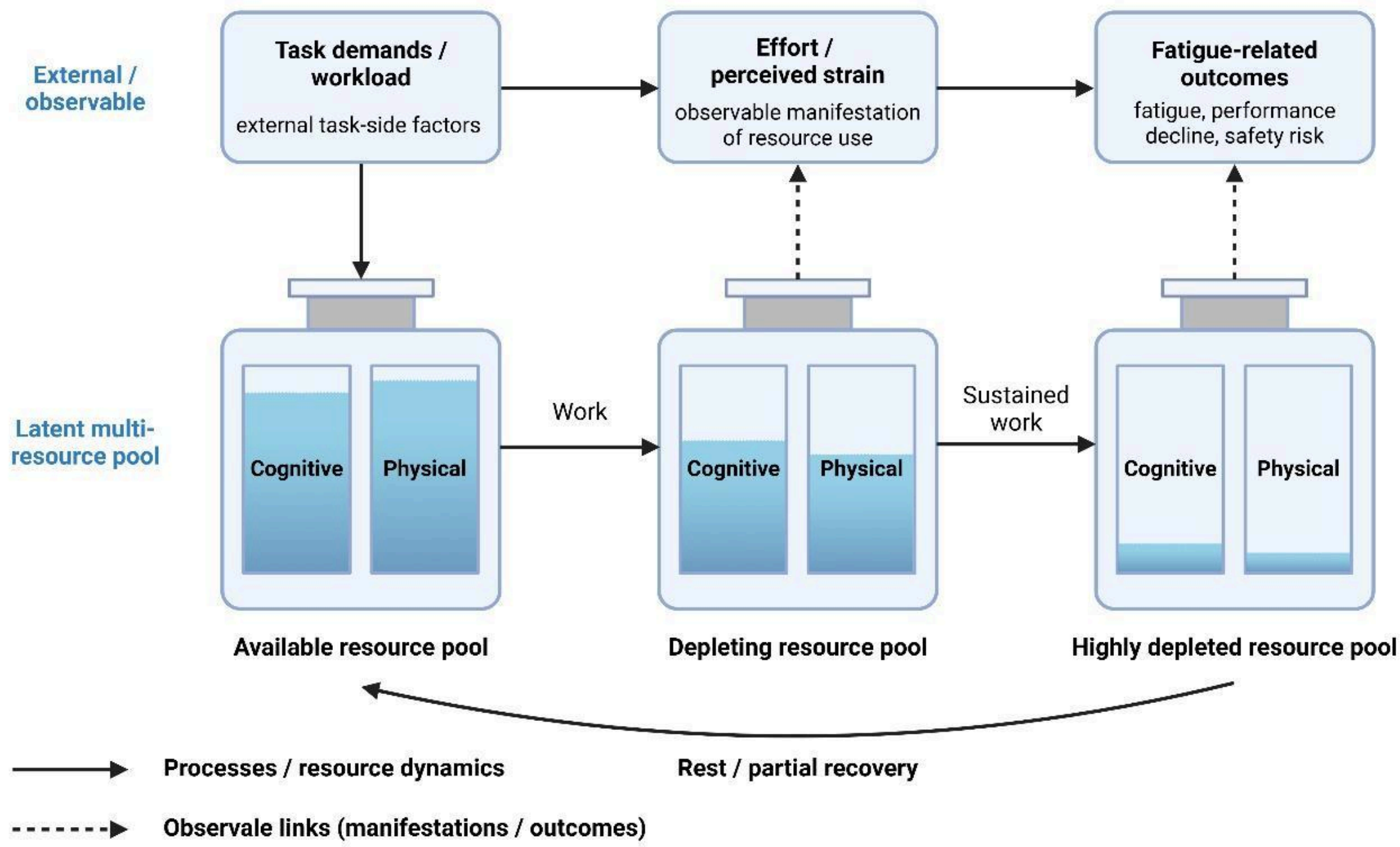

**Figure 1.** Conceptual relationship among workload, resource pool, perceived strain, fatigue, and recovery.

This distinction clarifies why perceived strain is used as the central worker-state variable in this study. Workload is task-defined and does not directly indicate how a worker experiences the task. Fatigue is a lagged outcome of sustained resource expenditure and therefore does not directly reveal how worker state evolves during work and rest [42]. Perceived strain occupies an intermediate position between workload and fatigue because it can be repeatedly observed during both work and rest. Therefore, perceived strain is used here as an interpretable, planning-oriented proxy for time-dependent worker state, rather than a physiological measure of fatigue.

### 2.4 Measurement approaches for perceived cognitive and physical strain

Human-state measurement in HRC can be based on physiological or self-reported measures. Physiological measures, including electroencephalography (EEG), functional near-infrared spectroscopy (fNIRS), electromyography (EMG), heart rate, heart-rate variability, skin conduction, and motion-based indicators, can provide valuable information about neural, muscular, cardiovascular, and behavioral responses [10,43–45]. Therefore, physiological sensing is an important approach for understanding and responding to human state in collaborative work, including workload classification, fatigue detection, safety-aware interaction, and real-time human-state monitoring.

However, physiological measures also require careful interpretation when used for planning-oriented perceived strain modeling. They often involve specialized sensors, signal preprocessing, artifact removal, feature extraction, and task-specific calibration [21,46]. More importantly, the relationship between physiological signals and perceived cognitive and physical effort, fatigue, and recovery may vary across individuals, tasks, and environments. Thus, physiological measures provide valuable objective evidence, but additional modeling is often needed before they can be translated into interpretable planning variables.

Self-reported scales provide a complementary approach that is well aligned with the objective of this study. Instruments such as Borg's Rating of Perceived Exertion [47], the Rating Scale for Mental Effort (RSME) [48], NASA-TLX [49], and the Paas mental effort scale [50] are widely used in human factors research to assess perceived exertion, mental effort, and workload. These scales are minimally intrusive, easy to administer repeatedly, and supported by prior validation. More importantly, these scales provide an interpretable representation of perceived strain dynamics within the studied task context [51]. Prior studies have successfully applied linear and nonlinear modeling of such measures to characterize effort accumulation and recovery processes [52,53]. For repeated work–rest experiments, this makes them suitable for characterizing temporal patterns of perceived strain accumulation and recovery.

Despite their widespread use, few studies have employed self-reported measures within HRC prefabrication scenarios to explicitly identify perceived strain accumulation and recovery dynamics under both cognitive and physical demands. Experimental protocols that align repeated self-report measurement with dynamic model identifiability remain lacking. This gap motivates the present study's focus on repeatedly measured perceived cognitive and physical strain as planning-oriented proxies for human state dynamics in prefabricated construction HRC.

## 2.5 Research Gaps and Objectives

Despite progress in incorporating human state into HRC planning, theoretical modeling of work–rest processes, and experimental effort measurement, existing work remains fragmented for planning-oriented modeling of perceived strain dynamics in HRC systems in prefabricated construction.

### *2.5.1. Research Gaps*

Two critical research gaps can be identified.

- First, the temporal dynamics of perceived cognitive and physical strain accumulation and recovery are not well characterized for planning-oriented applications with empirical evidence. Many HRC planning studies rely on static workload indicators or assumed functional forms without empirically examining how perceived strain evolves across working and resting phases. As a result, differences between cognitive and physical dynamics, phase-dependent behaviors, and conditional effects are poorly understood in prefabrication contexts, thereby constraining the empirical basis for future human-state-aware planning models.
- Second, current HRC experimental protocols in prefabricated construction are rarely designed to generate temporally structured data suitable for perceived strain modeling. Existing experimental studies often focus on factor effects, condition comparison, and fatigue detection rather than characterizing accumulation and recovery processes of perceived strain under both cognitive and physical demands. This limitation reduces the availability of empirical data needed to identify functional forms, estimate model parameters, and compare accumulation and recovery dynamics across cognitive and physical strain dimensions.

#### *2.5.2 Research Objectives*

To address these research gaps, this study develops an empirically grounded, planning-oriented approach for characterizing perceived cognitive and physical strain dynamics in prefabricated construction HRC through three interrelated objectives.

- First, to empirically characterize how cognitive and physical strain accumulate and recover over time during working and resting phases under different collaboration conditions. Functional forms are evaluated and mixed-effects modeling is used to separate experimental effects from inter-individual variability.
- Second, to develop interpretable perceived strain models that can inform future planning and scheduling research. These models translate observed perceived-strain trajectories into parameterized forms that may serve as candidate inputs for future work–rest design and task allocation.
- Third, to design and evaluate an experimental protocol aligned with perceived strain modeling needs, using repeated work–rest cycles and self-reported measurements. The protocol is abstracted into a framework that can guide future human-centric HRC experiments requiring temporally structured perceived human-state data.

## 3. METHODOLOGY

### 3.1 Experimental Protocol Design

Figure 2 provides an overview of the experimental protocol. A controlled laboratory experiment was designed to examine the temporal evolution of cognitive and physical strain during prefabricated construction manufacturing tasks performed with and without robotic collaboration. Each participant completed two experimental sessions within a single laboratory visit. Each session consisted of a 60 min working phase followed by a 30 min rest phase. This work–rest structure was designed as a feasible laboratory protocol to support observable perceived-strain accumulation and recovery, rather than to directly replicate an industry shift schedule. Two experimental conditions were considered, and all participants experienced both conditions. In the human-only condition, all task components were executed manually by the participant, with the robot present in the same workspace but not activated for task execution. In the HRC condition, a robotic arm assisted with selected repetitive or positioning-intensive operations within the same task workflow. The order of condition exposure was counterbalanced across participants, with half of the participants completing the human-only condition first and the remaining participants completing the HRC condition first. The human-only condition was included because, even in HRC workflows, some task steps may still be assigned to human-only execution due to task simplicity, robot reachability, tool-changing requirements, or resource constraints. This condition also allowed us to compare perceived strain dynamics between human-only and HRC conditions, as a prior study suggests that working with an active robot may affect human performance and responses [54].

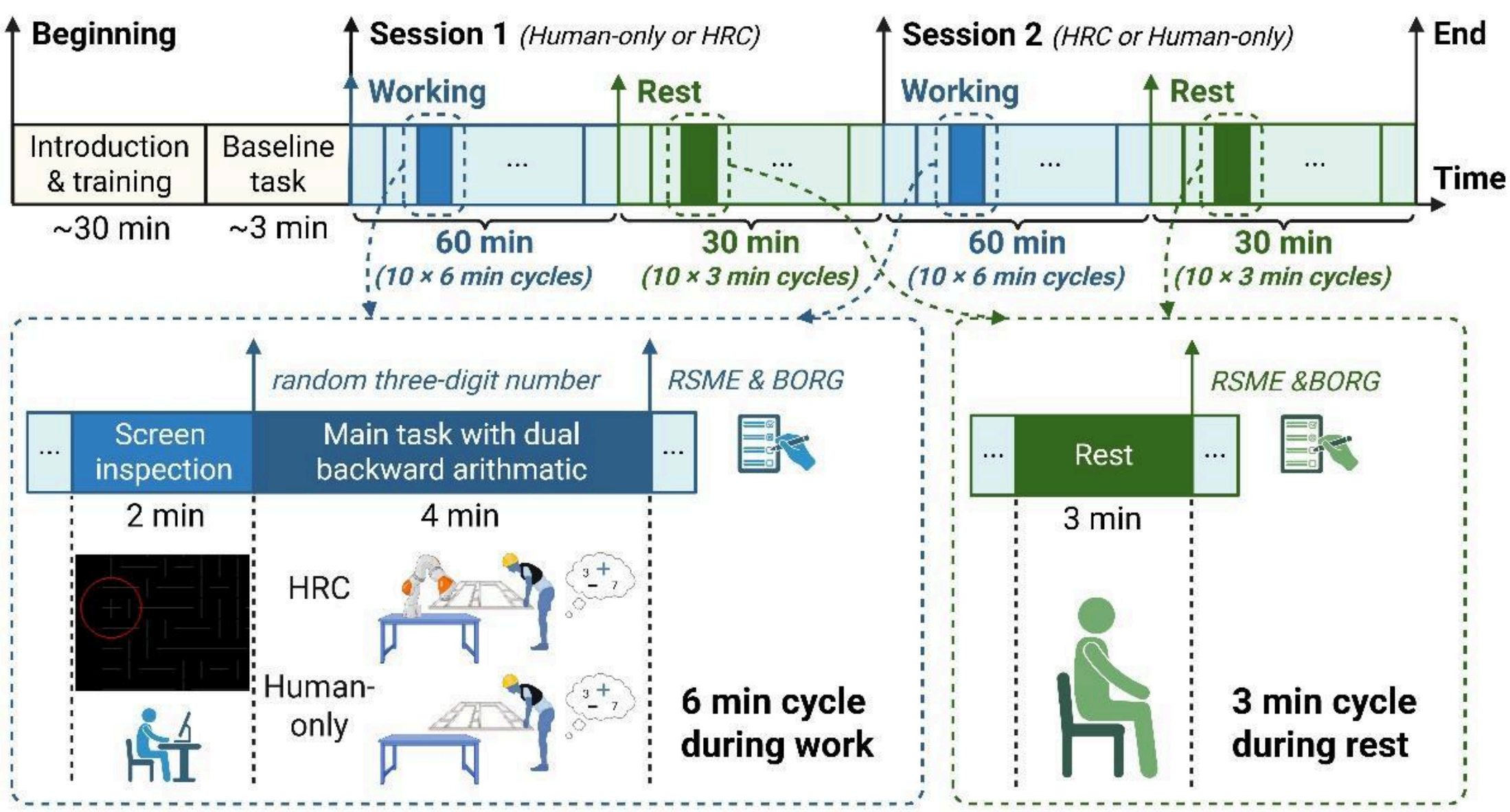

**Figure 2.** Overview of the experimental protocol, illustrating the two-session, within-subject counterbalanced design; phase structure (working and rest); task composition within working cycles (main prefabrication with cognitive dual task and screen inspection); and timing of perceived strain measurements.

During each working phase, participants completed ten repeated 6 min work cycles. Each cycle consisted of a 4 min primary prefabrication task performed continuously, as well as a 2 min screen-based inspection task. This cycle structure was designed to approximate frequent alternation between physical execution and cognitive monitoring commonly observed in prefabricated construction environments. Self-reported strain ratings were collected at predefined time points within each cycle, as detailed in Section 3.2. Between working phases, participants remained seated during the rest phases without task engagement. No additional cognitive or physical tasks were introduced during rest, and perceived strain ratings were collected periodically to characterize recovery dynamics following sustained work.

Prior to the formal experiment, participants received standardized instructions and completed a brief familiarization period to ensure understanding of the task procedures and overall experimental flow. A short baseline task, in which participants drove ten screws into a lumber board, was conducted to establish basic tool familiarity, and baseline perceived strain measurements were recorded before the first working phase to serve as reference levels for subsequent analyses.

### *3.1.1 Main Prefabrication Task*

The main prefabrication task was designed to approximate a typical wooden wall panel manufacturing workflow under controlled laboratory conditions. The task was based on an established experimental framework developed in our prior works [20,54]. The workflow was decomposed into nine sequential subtasks (T1–T9), including framing, panel handling, screwing, insulation insertion, and panel relocation, as illustrated in Figure 3.

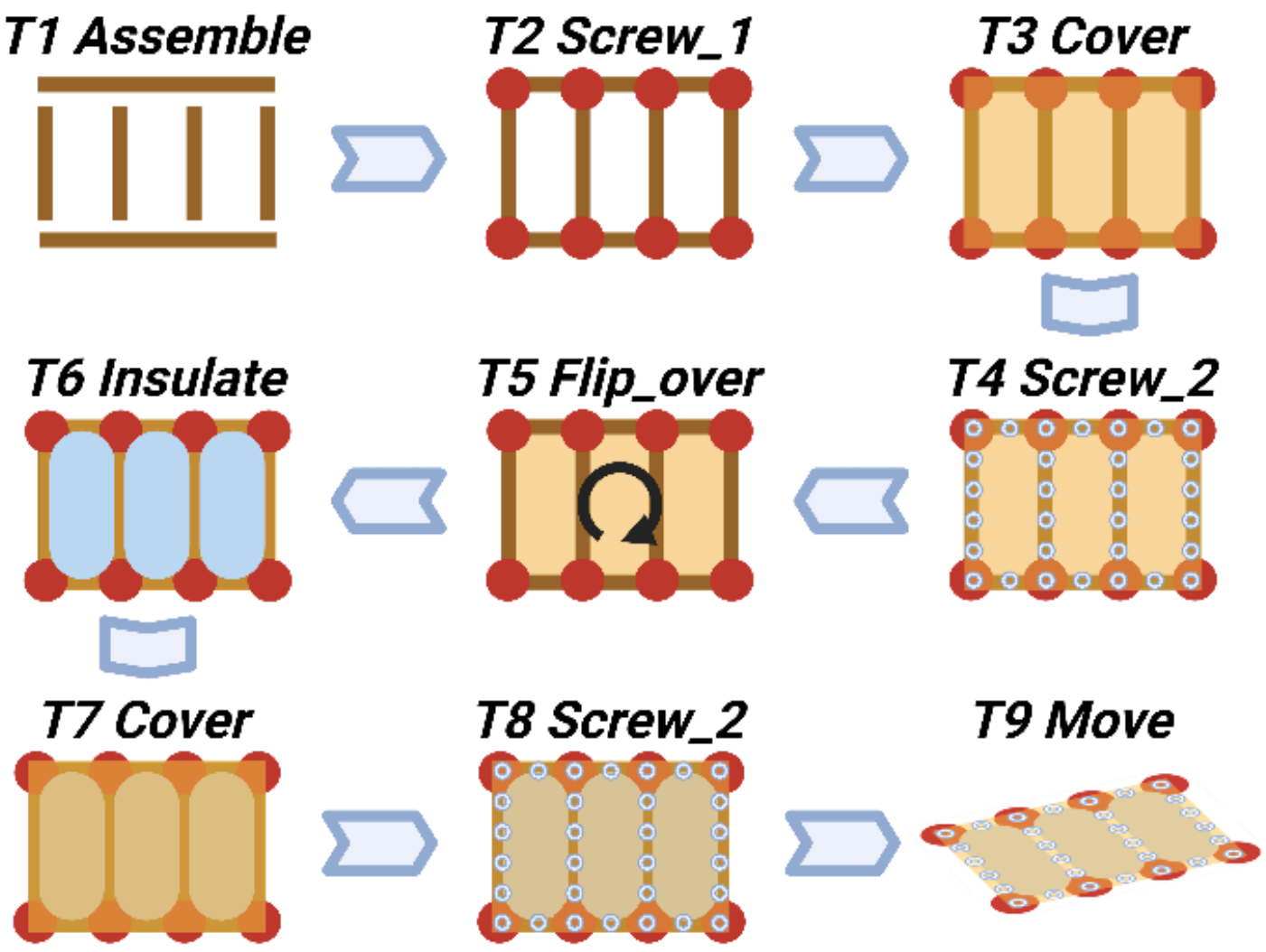

**Figure 3.** Workflow of the designed wooden wall panel manufacturing task [54]

During each 4 min working segment, participants were instructed to perform the manufacturing task continuously and as efficiently as possible. Task execution was interrupted only for scheduled perceived strain reporting and predefined task transitions. Across the full 60 min working phase, participants resumed the task from the point reached in the previous segment, thereby preserving task continuity and cumulative workload exposure over time.

### *3.1.2 Robotic Programming in the HRC Condition*

In the HRC condition, a UR10e robotic arm equipped with a Robotiq 2F-85 parallel gripper was integrated into the main prefabrication workflow. Robot motions were predefined within a workspace-level digital twin environment and executed using offline programming in RoboDK (version 5.7.0), ensuring consistent behavior and repeatability across participants and trials.

Robotic collaboration was implemented across multiple subtasks to reflect practical HRC scenarios, in which robots support positioning-intensive and repetitive actions while humans retain control over complementary operations, as illustrated in Figure 4. During subtask T1, the robot placed two long timber members, while the participant manually positioned the remaining four shorter members. In subtask T2, the robot performed screwing operations for the four screws closest to its reachable workspace, whereas the participant completed the four screws located at the distal end of the panel. A similar division of labor was applied in subtasks T4 and T8, where the robot executed screwing operations for the ten screws nearest to its reach, and the participant completed the remaining twenty screws.

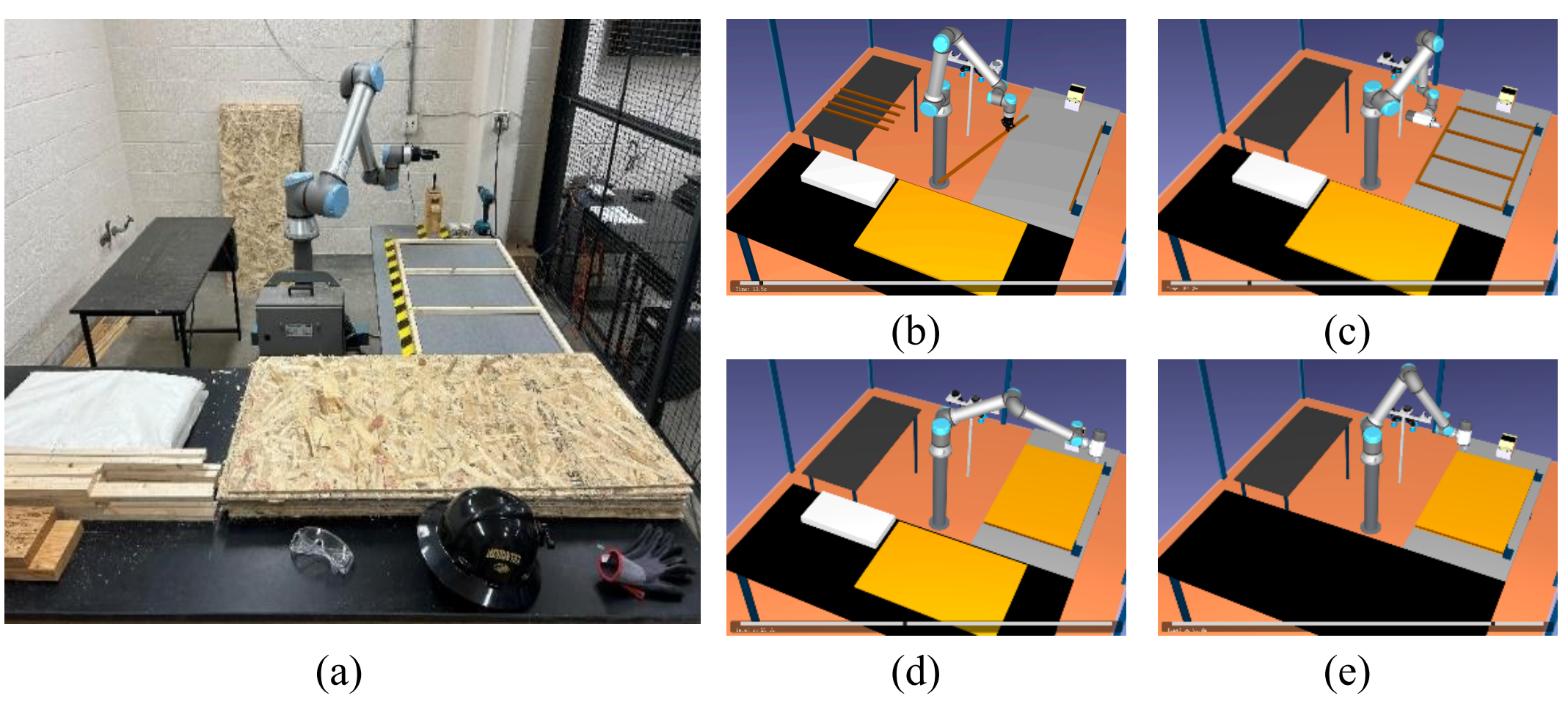

**Figure 4.** Laboratory workspace and robot execution simulated on the RoboDK platform within the prefabrication workflow: (a) laboratory workspace, (b) T1, (c) T2, (d) T4, and (e) T8

All remaining subtasks (T3, T5, T6, T7, and T9) within the workflow were performed solely by the participant due to laboratory hardware constraints. These subtasks involved simple manipulations such as flipping or repositioning panels and accounted for a minimal proportion of the total task duration, typically ranging from a few seconds to several tens of seconds.

#### *3.1.3 Cognitive Dual Task and Intermittent Screen Inspection*

To impose sustained cognitive demand during physical task execution, a backward arithmetic dual task was embedded within each 4 min working segment. Backward arithmetic tasks, such as serial subtraction, have been widely used in motor-cognitive dual-task paradigms to impose cognitive demand because they require working-memory updating, sustained attention, and continuous information processing [55]. At the start of each segment, participants were provided with a randomly generated three-digit number and instructed to repeatedly subtract seven. An auditory prompt was delivered every 12 s, and participants verbally reported the updated value following each prompt. This task was used to approximate task-relevant cognitive demands in prefabrication settings, such as tracking component dimensions, rather than to assess arithmetic proficiency.

In addition to the dual task, a 2 min screen inspection task was included at the start of each 6 min work cycle to represent intermittent monitoring and quality inspection commonly required in prefabricated construction. This task was implemented as a visual search task, which can impose cognitive demand by requiring sustained visual attention, target detection, and decision-making among distractors [56]. During this period, participants continuously performed the task and were instructed to maintain an appropriate balance between response rate and accuracy until the task terminated automatically at the end of the 2 min interval. Participants viewed a visual display and made a binary decision regarding the presence or absence of a target stimulus. As illustrated in Figure 5, the display consisted of a full-screen array of short horizontal or vertical line segments, with a “+” intersection appearing in a subset of trials. Participants pressed the left Shift key if an intersection was detected and the right Shift key otherwise. The task was implemented using Psychology Experiment Building Language (PEBL, version 2.1) [57], with stimulus presence and screen position randomized across trials.

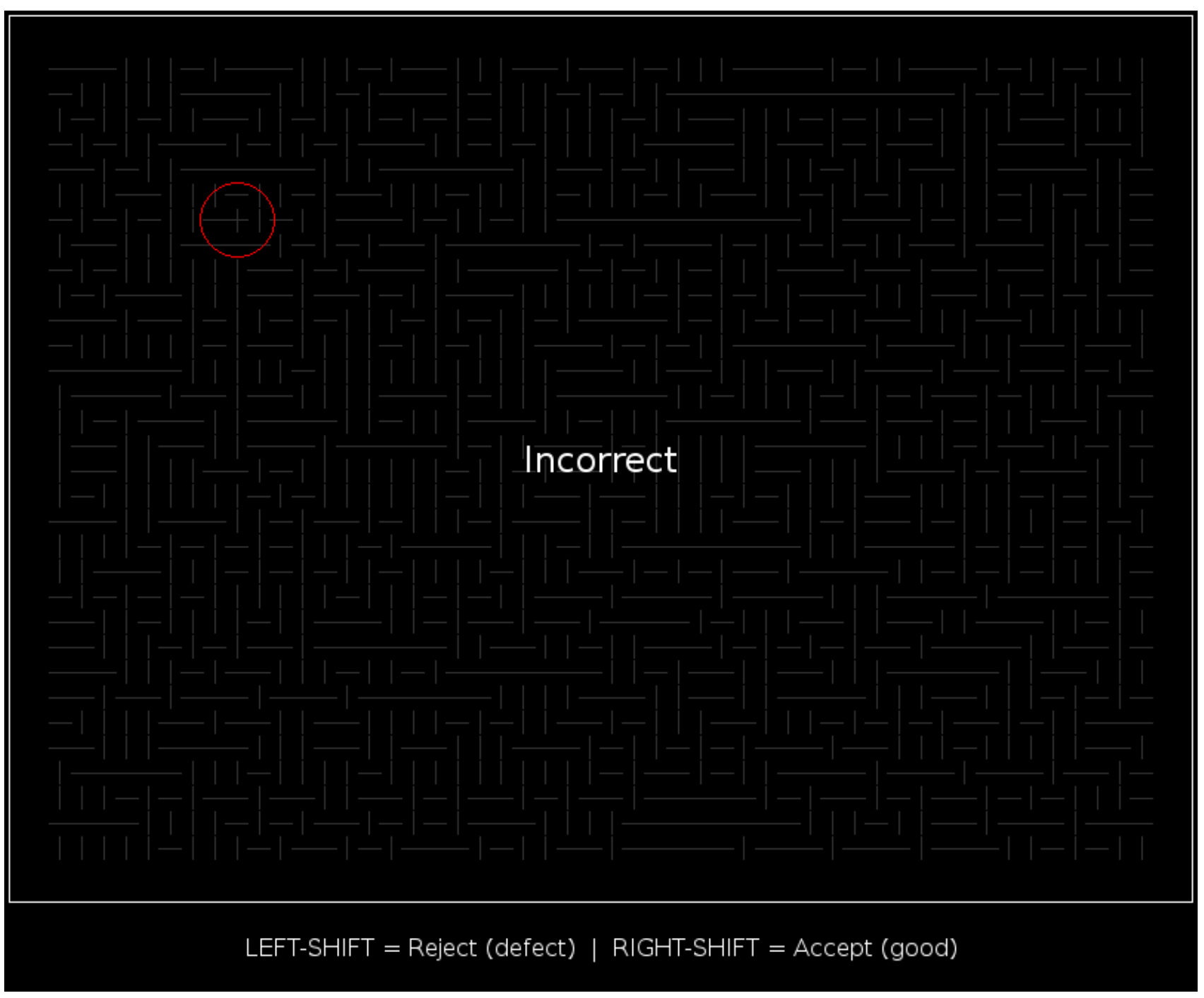

**Figure 5.** Example of the screen inspection task used during the experiment. Participants viewed arrays of short line segments and indicated whether a “+” intersection was present

**3.2 Participants and Data Collection**

A total of sixteen participants were recruited for this study, including four females and twelve males. All participants were adults (≥18 years old) and reported no musculoskeletal injuries or health concerns that could affect physical task execution. Participants were recruited through campus-wide announcements and direct email invitations. The experimental protocol was approved by the Institutional

Review Board (IRB) at Michigan Technological University, and all participants provided written informed consent prior to participation. Table 2 summarizes demographic details of participants as well as previous experience with woodworking tools and HRC.

**Table 2.** Participant demographic and experience summary

| Characteristic | Category / summary | n (%) |
|---|---|---|
| Age, years | Mean ± SD | 23.3 ± 5.8 |
| | Range | 19–38 |
| Gender | Male | 12 (75.0%) |
| | Female | 4 (25.0%) |
| Dominant hand | Right | 16 (100.0%) |
| Education level | Bachelor’s | 11 (68.8%) |
| | Master’s | 2 (12.5%) |
| | Ph.D. | 3 (18.8%) |
| Previous woodworking / electric screwdriver experience | A little | 5 (31.3%) |
| | Moderate | 7 (43.8%) |
| | Extensive | 4 (25.0%) |
| Previous HRC experience | No | 10 (62.5%) |
| | Yes | 6 (37.5%) |
| Known physical condition affecting arm/hand movement | No | 16 (100.0%) |
| Current or past back injury/pain affecting physical activity | No | 16 (100.0%) |

During the experiment, mental and physical strain were measured repeatedly using two established self-report scales. Cognitive strain was assessed using the Rating Scale for Mental Effort (RSME) with a scale from 0 to 150 [48], and physical strain was assessed using Borg’s Rating of Perceived Exertion (BORG) with a scale from 6 to 20 [47]. Both scales were collected at regular intervals during both working and rest phases, with measurements obtained every 6 min during working periods and every 3 min during rest periods, yielding dense time-series observations for each participant across experimental conditions and sessions. Both scales were administered digitally to ensure consistent presentation, as shown in Figure 6. The resulting dataset has a repeated-measures structure, with multiple time-indexed observations nested within participants across task phases, experimental conditions, and sessions.

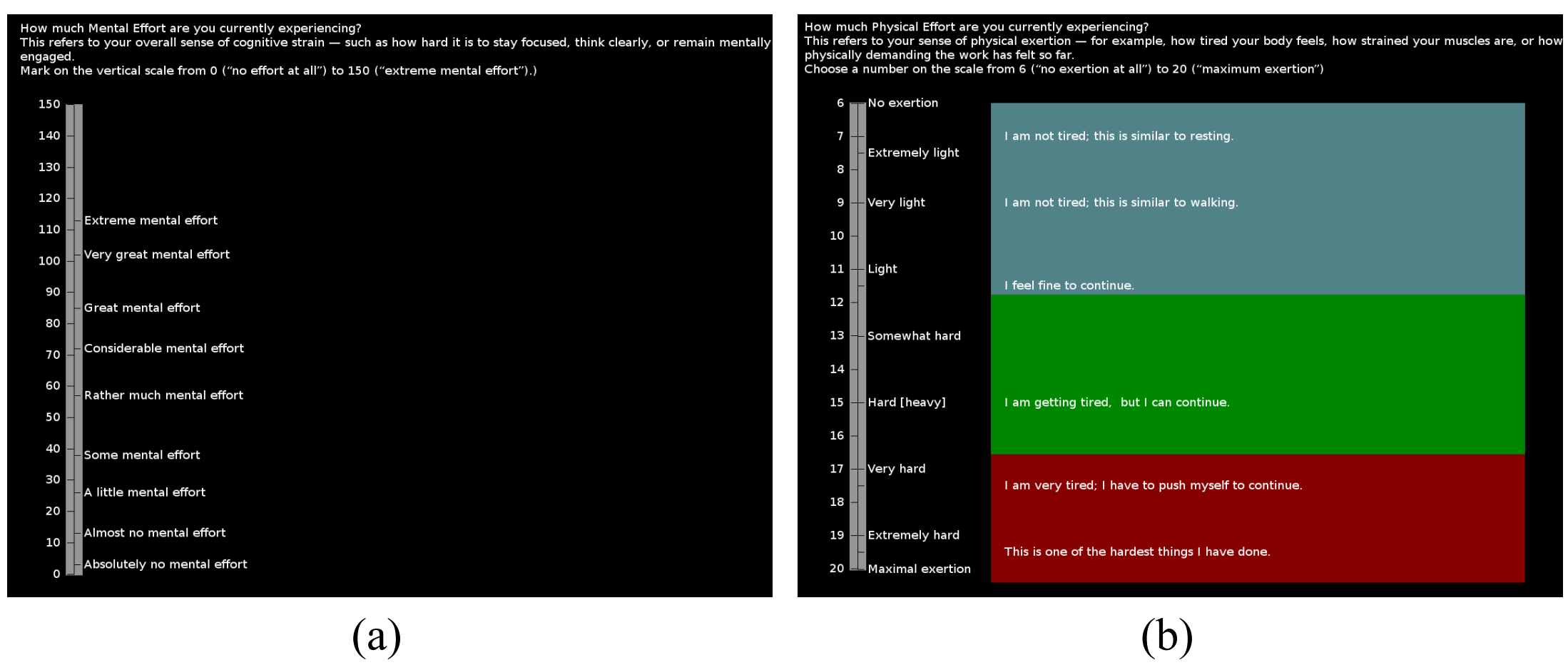

**Figure 6.** Digital presentation of self-report scales used in the experiment: (a) Rating Scale for Mental Effort (RSME) and (b) Borg Rating of Perceived Exertion (BORG)

### 3.3 Data Analysis

This study employed a structured, multi-stage data analysis strategy to examine the temporal dynamics of cognitive and physical strain, progressing from data preprocessing and subject-level functional-form exploration to model comparison, population-level examination, and mixed-effects model development, as illustrated in Figure 7. All statistical analyses were conducted in R (version 4.4.2) [58].

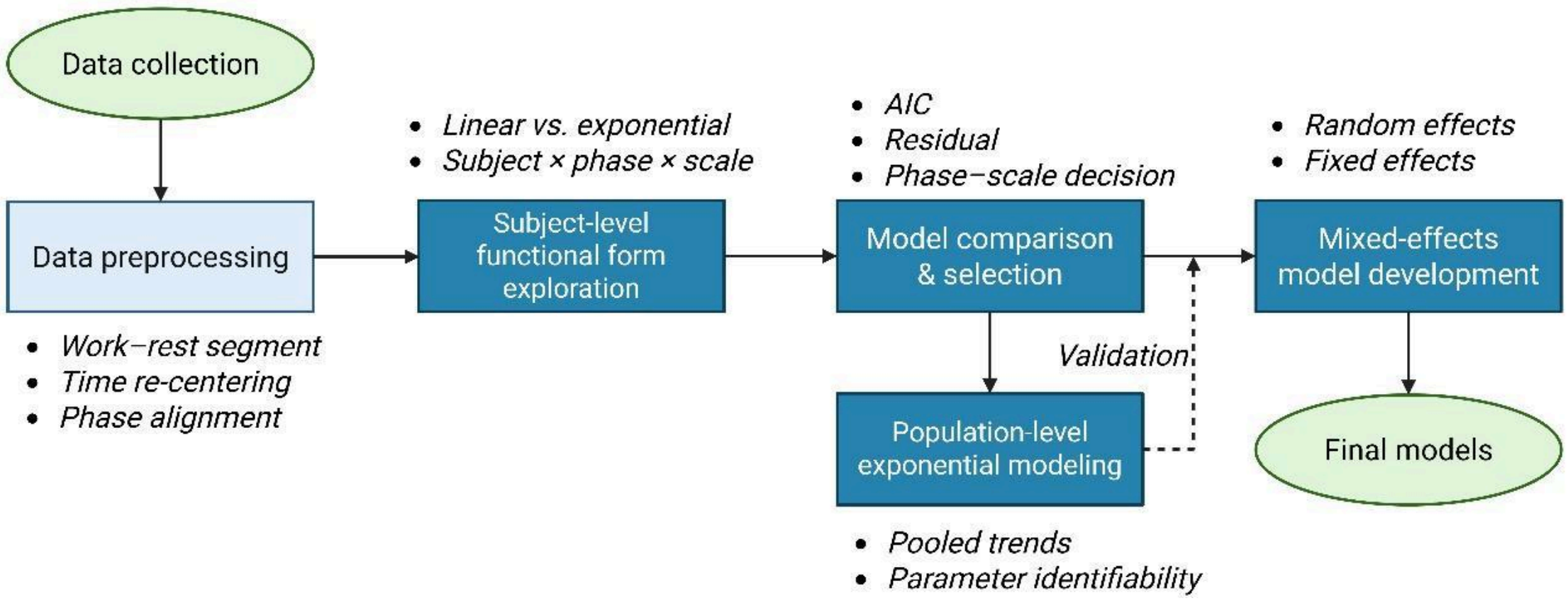

**Figure 7.** Overview of the multi-stage data analysis workflow, illustrating preprocessing, functional-form exploration, model selection, population-level modeling, and mixed-effects model development

#### *3.3.1 Data Preprocessing*

Experimental data were segmented according to the predefined work–rest protocol, consisting of two repeated blocks of 60 min working followed by 30 min rest. Four analysis segments were defined: Session 1 work, Session 1 rest, Session 2 work, and Session 2 rest. For each segment, a relative time variable (segment_time) was constructed to represent elapsed time since the start of that segment. For rest segments, the terminal observation of the immediately preceding working period was explicitly treated as the starting point (segment_time = 0) of recovery. This anchor ensured continuity between accumulation and recovery processes and enabled direct modeling of rest-phase dynamics from a known initial state.

#### *3.3.2 Candidate Functional Form Fitting and Comparison*

To characterize the temporal evolution of cognitive and physical strain within each analysis segment, two candidate functional forms were evaluated: a linear model and a nonlinear exponential model. These alternatives were selected to assess whether strain dynamics exhibited approximately linear trends over time or nonlinear patterns consistent with accumulation and recovery processes.

For working phases, the exponential model was specified to represent monotonic accumulation toward an asymptotic level (Eq. (1)), where $a$ denotes the initial strain level at the start of the segment, $b$ represents the accumulation magnitude, and $k$ is the accumulation rate.

$$y(t) = a + b(1 - e^{-kt}) \qquad \text{Eq. (1)}$$

For rest phases, an exponential decay model was specified to represent recovery toward an asymptotic level (Eq. (2)), where a corresponds to the recovered asymptotic strain level, $a + b$ represents the strain level at the onset of rest, and $k$ captures the recovery rate.

$$y(t) = a + be^{-kt} \quad \text{Eq. (2)}$$

In parallel, linear model (Eq. (3)) was included as parsimonious alternatives for both working and rest phases.

$$y(t) = \beta_0 + \beta_1 t \quad \text{Eq. (3)}$$

Both linear and nonlinear models were fitted at the subject × scale × segment level to preserve individual temporal patterns and avoid aggregation effects. Nonlinear models were estimated using nonlinear least squares with bounded optimization and multiple starting values to improve numerical stability.

Model comparison was conducted using the Akaike Information Criterion (AIC) and inspection of residual patterns over segment time. Differences in AIC were used to assess relative model adequacy, while residual diagnostics were examined to identify systematic temporal structure indicative of model misspecification. The results of these comparisons were used to inform subsequent modeling decisions regarding appropriate functional forms for each phase–scale combination.

***3.3.3 Population-Level Exponential Modeling***

Following the subject-level functional form evaluation, population-level exponential models were fitted as an exploratory step to summarize average accumulation and recovery trends for cognitive (RSME) and physical (BORG) strain. This analysis aimed to assess the identifiability and interpretability of key model parameters when individual-level variability was not explicitly modeled.

Data were pooled across participants within each scale (RSME, BORG) and phase (working, rest). Exponential models were fitted separately for each scale–phase combination using nonlinear least squares. Consistent with the subject-level specification, working-phase dynamics were modeled using the exponential accumulation formulation (Eq. (1)), whereas rest-phase dynamics were modeled using the exponential recovery formulation (Eq. (2)). Model estimation employed bounded nonlinear least squares with multiple starting values to enhance numerical stability. Model adequacy at this stage was evaluated descriptively using fitted trajectories, parameter estimates, and residual patterns.

#### *3.3.4 Mixed-Effects Modeling Strategy*

Mixed-effects modeling was employed as the primary inferential framework to quantify experimental effects while accounting for repeated measurements and pronounced inter-individual variability. Analyses were conducted separately by phase (working vs. rest) and scale (RSME vs. BORG), consistent with the phase-dependent functional forms identified in the preceding analyses and to avoid conflating accumulation and recovery processes within a single model.

Model development followed a hierarchical, stepwise strategy emphasizing identifiability and parsimony. For each phase–scale combination, a baseline model was first specified using the supported functional form (Sections 3.3.2–3.3.3) without random or fixed effects. Model complexity was then increased progressively, first by introducing random effects to capture between-subject variability in baseline levels and temporal dynamics, and subsequently by adding fixed effects representing hypothesized influences of experimental Condition and Session.

Model extensions were evaluated using likelihood ratio tests (LRTs) under maximum-likelihood estimation, information criteria including the AIC and Bayesian Information Criterion (BIC), convergence behavior, parameter plausibility, and residual diagnostics. Because multiple LRTs were performed as part of a hierarchical model-building process rather than as independent hypothesis tests, LRT results were interpreted together with AIC/BIC, convergence behavior, parameter plausibility, and residual diagnostics rather than used as the sole selection criterion. When candidate specifications resulted in convergence failures or non-identifiable parameters, simpler models were retained to ensure stable and interpretable inference.

For each phase–scale combination, the final model was defined as the most parsimonious specification that achieved stable estimation, adequately captured between-subject variability, and supported experimental fixed effects where identifiable. To improve reporting transparency, key fixed-effect estimates from the final selected models will be reported with 95% confidence intervals in Supplementary Table S1. Estimated effects were then synthesized across phases and strain dimensions in the Discussion (Section 5.2) to facilitate interpretation of accumulation and recovery mechanisms.

# 4. RESULTS

## 4.1 Functional Form Comparison of Strain Dynamics

Functional form comparisons revealed clear phase-dependent differences in strain dynamics for both cognitive (RSME) and physical (Borg) measures. Linear and exponential models were evaluated at the subject level for each task segment, and their relative performance was assessed using information criteria and residual diagnostics.

### *4.1.1 Information-Criteria Results: Nonlinear Recovery Is Strongly Supported*

As summarized in Table 3, exponential models consistently outperformed linear models during rest segments for both scales. Across both rest periods (S1-Rest and S2-Rest), median ΔAIC values were large and positive, indicating substantial improvements in model fit when exponential functions were used to represent recovery dynamics. For RSME, the exponential model was favored in nearly all subject-level fits during rest, while BORG ratings showed similarly strong support, with more than 80% of trajectories favoring the exponential form. In contrast, working segments (S1-Work and S2-Work) exhibited no consistent group-level advantage of either model. Median ΔAIC values during working phases were close to zero for both scales, and a substantial proportion of subject-level fits fell within the range of comparable performance ($|\Delta AIC| < 2$), indicating generally small differences in goodness-of-fit between linear and exponential models, rather than clear dominance of a single functional form. Notably, the proportion of fits favoring the linear model decreased from S1-Work to S2-Work for both Borg (0.375 to 0.25) and RSME (0.1875 to 0), suggesting a modest shift away from linear superiority in the second working period.

**Table 3.** Subject-level functional form comparison of linear and exponential models across task segments for cognitive (RSME) and physical (BORG) strain measures

| Scale | Segment | Median ΔAIC ($AIC_{Linear}$ - $AIC_{Exponential}$) | Proportion favoring exponential ($\Delta AIC \geq 2$) | Proportion favoring linear ($\Delta AIC \leq -2$) | Proportion with comparable fit ($|\Delta AIC| < 2$) |
|---|---|---|---|---|---|
| BORG | S1-Work | -1.326 | 0.125 | 0.375 | 0.5 |
| BORG | S1-Rest | 24.772 | 0.875 | 0.125 | 0 |
| BORG | S2-Work | 1.066 | 0.4375 | 0.25 | 0.3125 |
| BORG | S2-Rest | 25.023 | 0.8125 | 0.125 | 0.0625 |
| RSME | S1-Work | -0.101 | 0.3125 | 0.1875 | 0.5 |
| RSME | S1-Rest | 27.763 | 1 | 0 | 0 |
| RSME | S2-Work | -0.567 | 0.25 | 0 | 0.75 |
| RSME | S2-Rest | 30.535 | 0.9375 | 0 | 0.0625 |

#### *4.1.2 Residual Diagnostics: Recovery Shows Systematic Curvature under Linear Models*

Residual diagnostics further elucidated these phase-dependent differences as shown in Figure 8. During rest segments, residuals from linear models displayed pronounced systematic temporal structure, characterized by curvature and time-dependent bias, indicating an inadequate representation of recovery processes. In comparison, exponential models yielded residuals that were more symmetrically distributed around zero and exhibited minimal temporal trends, supporting their suitability for capturing nonlinear recovery dynamics. During working segments, residual patterns were broadly similar for linear and exponential models in S1-Work, with no strong evidence of systematic curvature uniquely favoring either functional form. In S2-Work, residuals from exponential models tended to be more symmetrically distributed around zero across both mental and physical strain measures, suggesting a modest improvement in capturing late-stage accumulation trends.

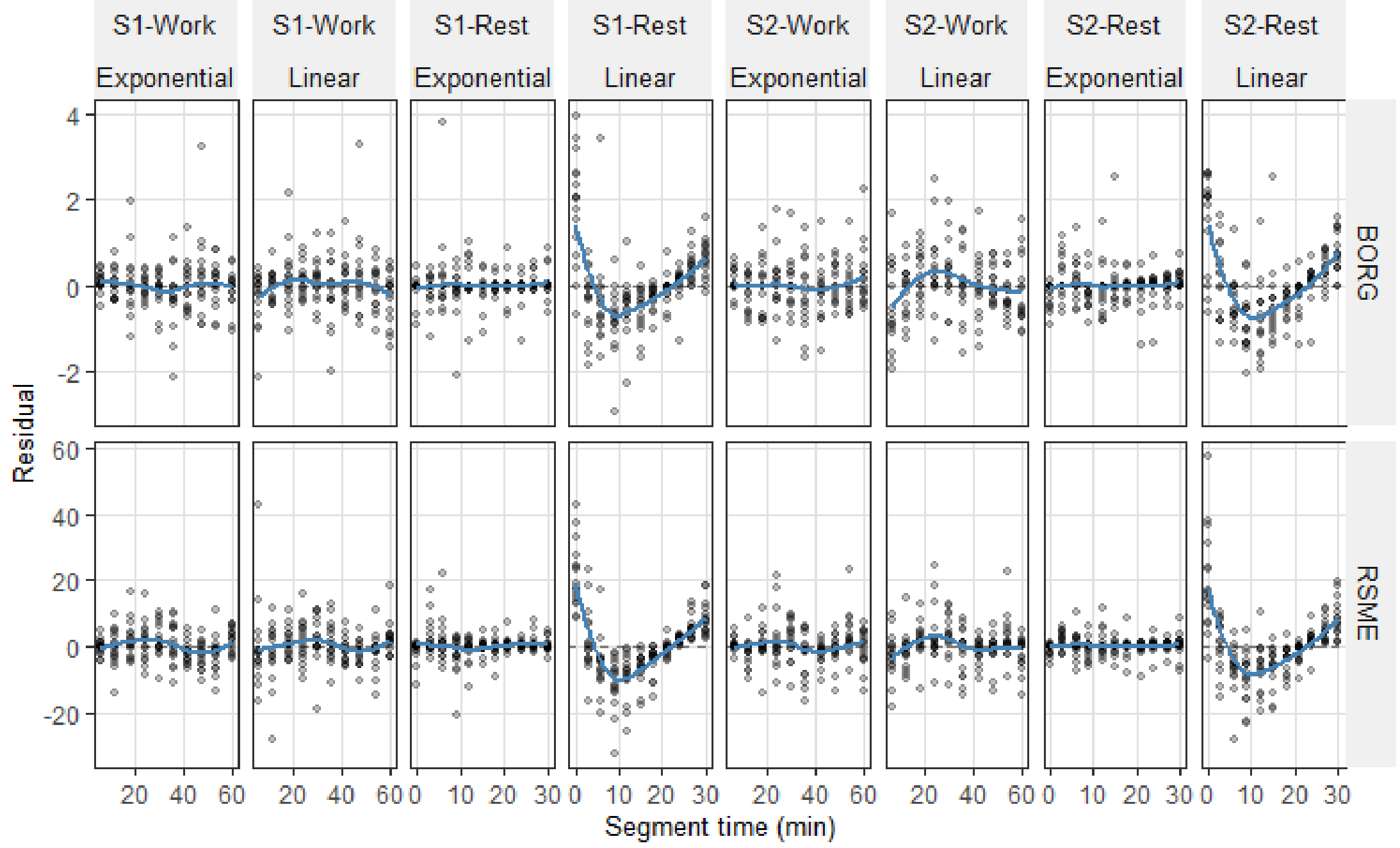

**Figure 8.** Residual diagnostics for linear and exponential models across task segments for cognitive (RSME) and physical (BORG) strain measures. Blue curves indicate locally estimated scatterplot smoothing (LOESS) trends for visual inspection of systematic temporal structure

Overall, the functional-form comparison suggests that recovery and accumulation should not be treated as the same temporal process. Recovery during rest showed strong nonlinear decay across both cognitive and physical measures, whereas accumulation during work did not show a single clearly dominant functional form. Therefore, both linear and exponential forms were retained as candidates for working phase modeling, with final selection based on parameter identifiability, stability under mixed-effects formulations, and interpretability rather than by goodness-of-fit alone.

## 4.2 Population-Level Exponential Modeling of Strain Dynamics

Based on the functional form selection results in Section 4.1, exponential models were fitted at the population level to characterize aggregate accumulation and recovery dynamics for cognitive (RSME) and physical (BORG) strain during working and rest phases.

### *4.2.1 Population-Level Fits: Recovery Is Rapid but Accumulation Is Gradual*

Population-level exponential fits revealed qualitatively distinct dynamics between working and rest phases, as shown in Figure 9. During rest periods, both RSME and BORG exhibited rapid decreases in strain over time, consistent with nonlinear recovery processes. In contrast, working phases showed more gradual increases in strain, with flatter trajectories and weaker curvature, indicating slower accumulation at the population level. This slow and weakly curved accumulation pattern is consistent with the subject-level results in Section 4.1, where linear and exponential models exhibited comparable goodness-of-fit during working segments.

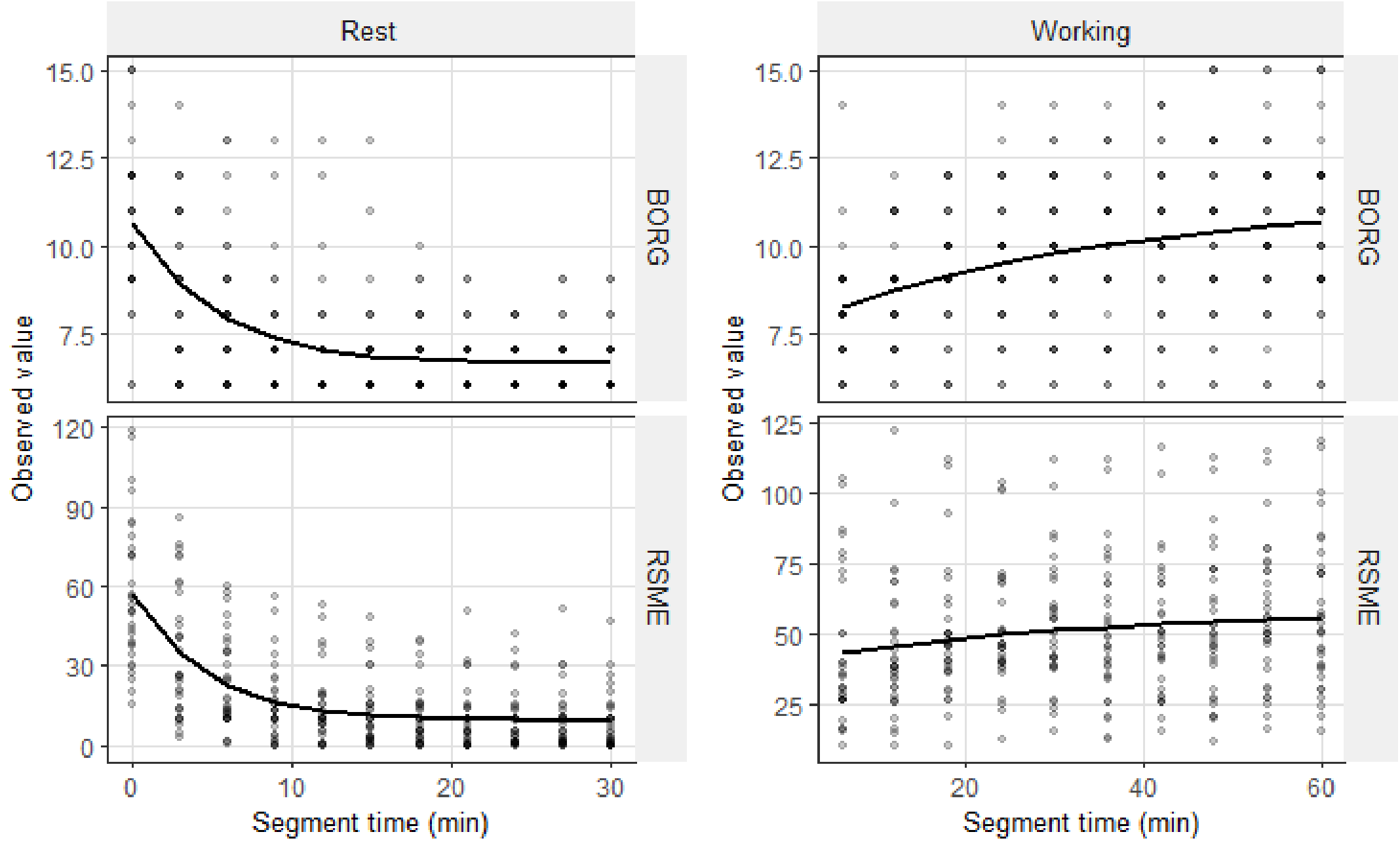

**Figure 9.** Population-level exponential fits for cognitive (RSME) and physical (BORG) strain during working and rest phases. Points represent observed measurements pooled across subjects, and solid lines indicate fitted population-level exponential trajectories

#### *4.2.2 Parameter Estimates and Residuals: Individual Differences Remain Evident*

Parameter estimates from the population-level exponential models are summarized in Table 4. Baseline ($a$) and amplitude ($b$) parameters were generally well estimated in both working and rest phases for cognitive (RSME) and physical (BORG) strain (p-value < 0.05). In contrast, the rate parameter ($k$) exhibited strong phase dependence. During rest phases, k was large and highly significant for both scales (p-value < 0.001), indicating rapid recovery dynamics. During working phases, however, $k$ estimates were small and not statistically significant for either RSME or BORG (p-value > 0.05), reflecting slower accumulation dynamics and limited identifiability of a single population-level accumulation rate.

**Table 4.** Estimated parameters of the population-level exponential fits for cognitive (RSME) and physical (BORG) strain during working and rest phases

| Scale | Phase | Parameter | Estimate | Std. error | P-value |
|---|---|---|---|---|---|

| BORG | Rest | a | 6.616 | 0.137 | < 0.001 |
|---|---|---|---|---|---|
| BORG | Rest | b | 4.013 | 0.277 | < 0.001 |
| BORG | Rest | k | 0.191 | 0.031 | < 0.001 |
| BORG | Working | a | 7.649 | 0.589 | < 0.001 |
| BORG | Working | b | 3.606 | 0.752 | < 0.001 |
| BORG | Working | k | 0.030 | 0.021 | 0.157 |
| RSME | Rest | a | 9.345 | 1.303 | < 0.001 |
| RSME | Rest | b | 47.474 | 2.830 | < 0.001 |
| RSME | Rest | k | 0.210 | 0.029 | < 0.001 |
| RSME | Working | a | 39.837 | 6.978 | < 0.001 |
| RSME | Working | b | 18.613 | 8.227 | 0.024 |
| RSME | Working | k | 0.031 | 0.048 | 0.517 |

Residual diagnostics provide additional insight into the limitations of population-level exponential modeling in the presence of substantial between-subject heterogeneity. As shown in Figure 10, residuals exhibited systematic subject-specific offsets in both working and rest phases, with residuals clustering consistently above or below zero for different individuals. This pattern indicates pronounced between-subject variability in strain dynamics that is not captured by a single population-level trajectory.

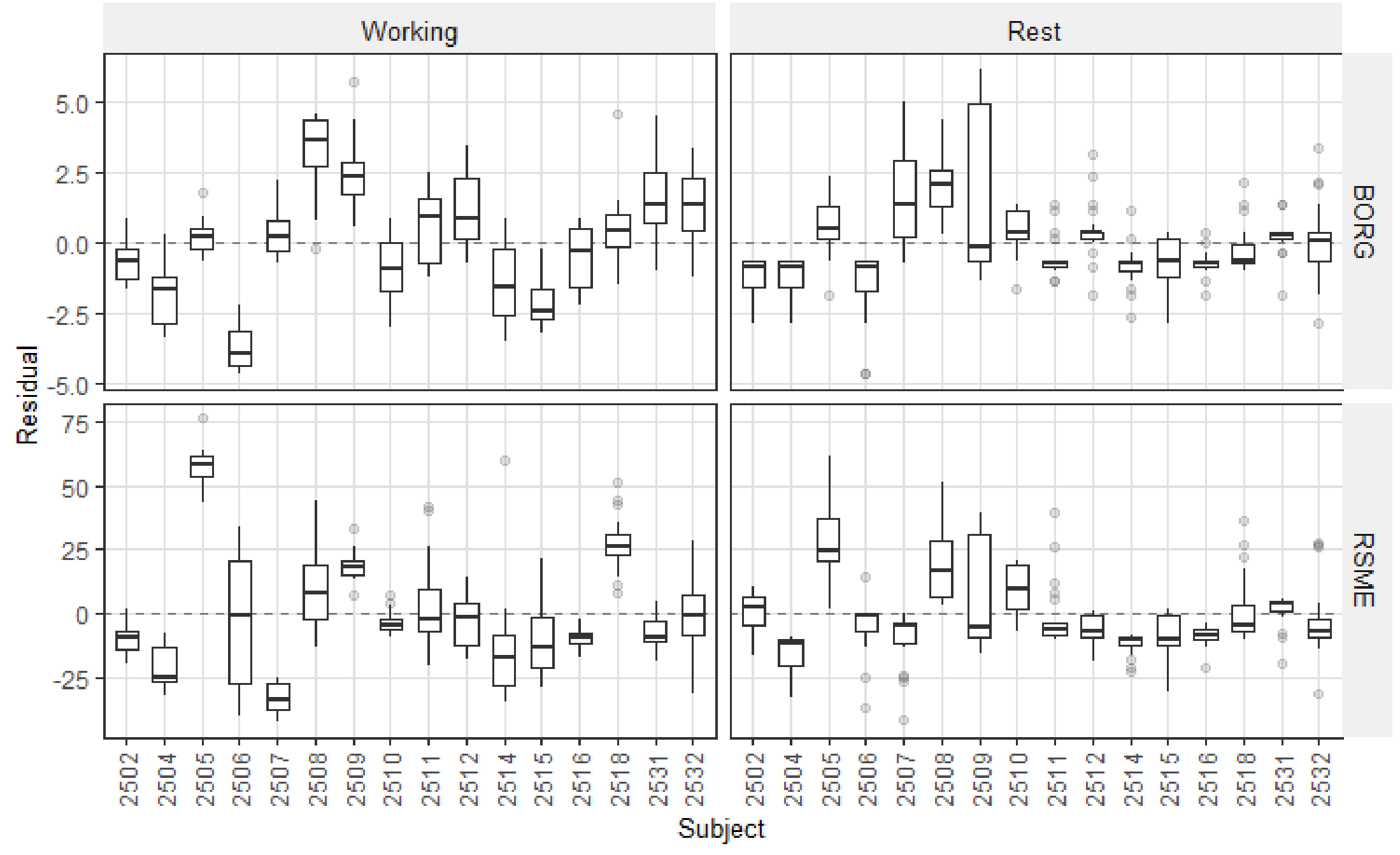

**Figure 10.** Residual distributions from population-level exponential models by subject for cognitive (RSME) and physical (BORG) strain during working and rest phases

The population-level results should therefore be interpreted as descriptive rather than sufficient for explaining worker-state dynamics. Population-level exponential models offer a useful descriptive summary of average accumulation and recovery trends, but the residual patterns show that meaningful individual differences remain after fitting a single population-level trajectory. The weak identification of the rate parameter ($k$) during working phases further suggests that mixed-effects modeling is needed before interpreting condition or session effects.

### 4.3 Mixed-Effects Modeling of Strain Dynamics: Separating Individual Variability and Experimental Effects

This section applied mixed-effects modeling to quantify between-subject variability (random effects) and the influence of experimental conditions (fixed effects) on strain dynamics while maintaining parameter identifiability. Physical (BORG) and cognitive (RSME) strain were analyzed separately for working and rest phases to account for their distinct accumulation and recovery characteristics.

#### *4.3.1 Working Phase: Accumulation Patterns During Task Execution*

For the working phase, exponential mixed-effects models were first evaluated to determine whether nonlinear accumulation dynamics were sufficiently identifiable to support subsequent inference on random and fixed effects. Given the weak identifiability of accumulation rate parameters observed in Section 4.2, the rate parameter k was fixed to scale-specific population-level estimates (BORG: $k$ = 0.03023 $min^{-1}$; RSME: $k$ = 0.03128 $min^{-1}$) to stabilize estimation and avoid overparameterization. Time was centered within each working episode (t = 0 at episode onset), such that model intercepts represent initial strain levels at the start of work.

4.3.1.1 Physical Strain Accumulation: Individual Variability Dominates

For working-phase physical strain (BORG), incorporating subject-level random effects substantially improved model fit, as shown in Table 5. LRTs provided strong evidence for between-subject variability in both the baseline parameter ($a$) and the accumulation amplitude ($b$). Allowing both parameters to vary by subject (model M3) yielded a markedly better fit than models with random effects on either parameter alone ($\Delta$AIC = 61–73; p-value < 0.001). The estimated correlation between random effects on $a$ and $b$ was weak ($\rho$ = 0.188), indicating largely independent inter-individual differences in baseline effort and accumulation magnitude.

**Table 5.** Random effects structure selection for working-phase physical strain (BORG) using exponential nonlinear mixed-effects models (rate parameter k fixed)

| Model | Random effects | Fixed effects | AIC | BIC | Log-likelihood | Test | LRT | P-value |
|---|---|---|---|---|---|---|---|---|
| M0 | None | None | 1372 | 1383 | -683 | M3 vs. M0 | 410.71 | < 0.001 |
| M1 | a | None | 1041 | 1056 | -516 | M3 vs. M1 | 76.20 | < 0.001 |
| M2 | b | None | 1029 | 1044 | -510 | M3 vs. M2 | 64.45 | < 0.001 |
| M3 | a, b | None | 968 | 991 | -478 | — | — | — |

Note: Attempts to incorporate fixed effects of Condition or Session on parameters $a$ or $b$ (Models M4–M7) did not converge to numerically stable solutions under maximum-likelihood estimation and are therefore not reported.

As illustrated in Figure 11, the selected nonlinear mixed-effects specification (M3) substantially reduced systematic subject-specific residual offsets relative to the population-level model, yielding residuals that were more symmetrically distributed around zero across individuals. This improvement supports M3 as a parsimonious and stable baseline for subsequent fixed-effects testing.

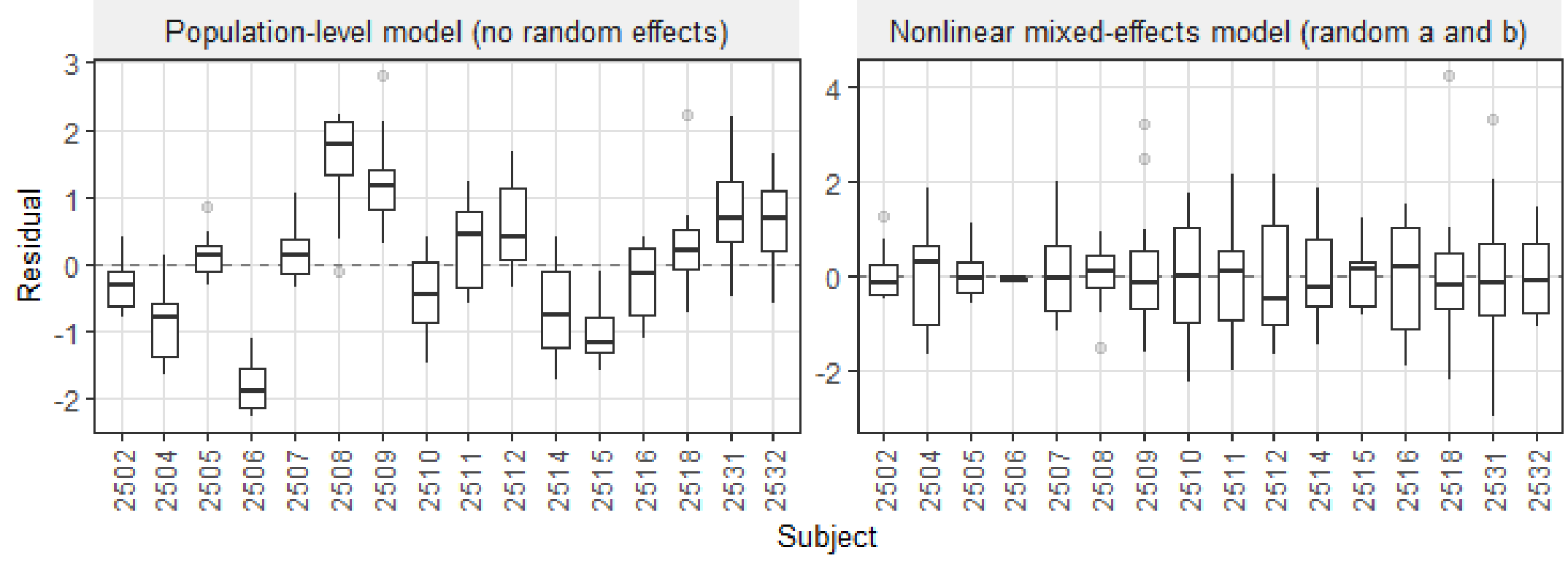

**Figure 11.** Residual distributions by subject for working-phase physical strain (BORG), comparing the population-level exponential model and the nonlinear mixed-effects model with subject-specific random effects

Fixed effects of experimental Condition and Session were then evaluated by extending the baseline model to allow these factors to influence $a$ or $b$ (models M4–M7). None of the condition- or session-augmented models converged to stable solutions. This indicates that, after accounting for subject-specific nonlinear accumulation dynamics, the working-phase BORG data do not support reliable identification of additional fixed effects within the nonlinear framework. This lack of identifiability persisted despite fixing the accumulation rate parameter, suggesting that the limited information content of the working-phase trajectories constrains inference on higher-level experimental effects.

In practical terms, this result suggests that physical strain accumulation during work was better explained by individual differences than by the tested collaboration condition or session.

4.3.1.2 Cognitive Strain Accumulation: Linear Growth with Condition and Session Effects

In contrast to physical strain, population-level exponential mixed-effects modeling for working-phase cognitive strain did not yield stable or identifiable fits, even in the absence of random or fixed effects. This suggests that perceived cognitive strain did not enter a reliably identifiable nonlinear accumulation regime within the one-hour working duration. Linear mixed-effects models were therefore adopted to characterize time-dependent RSME trajectories and to evaluate experimental effects. Figure 12 shows individual and mean working-phase RSME trajectories across sessions and conditions. Trajectories exhibited limited curvature and substantial inter-individual variability, supporting the use of linear mixed-effects models with subject-specific random effects.

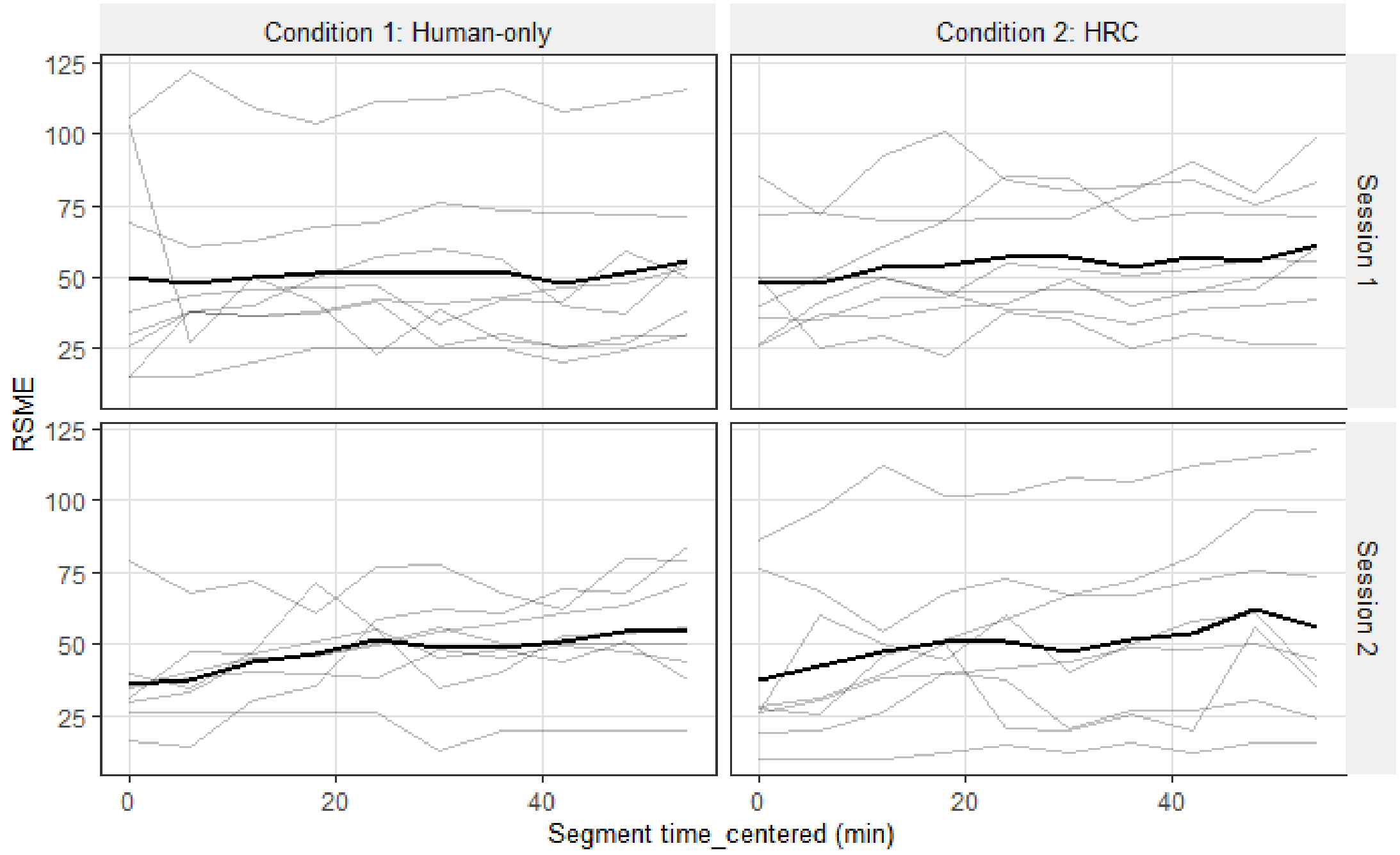

**Figure 12.** Individual working-phase RSME trajectories during Session 1 and Session 2 under Human-only and HRC conditions. Thin lines represent subject-level trajectories (time-centered within each working episode), and bold lines indicate the mean RSME trajectory within each session–condition combination

Model selection results are summarized in Table 6. Allowing subject-specific random slopes for time significantly improved model fit relative to a random-intercept-only specification (M1 vs. M0; LRT = 26.87, p-value < 0.001), indicating pronounced inter-individual variability in both baseline cognitive strain and temporal evolution during work. The selected random-effects structure included correlated random intercepts and slopes, with a modest negative correlation ($\rho = -0.19$), suggesting that participants with higher initial RSME tended to exhibit slightly slower increases over time.

**Table 6.** Model selection for working-phase cognitive strain (RSME) using linear mixed-effects models, models M0–M1 evaluate random-effects structures, whereas models M2–M5 assess Condition and Session fixed effects on the selected baseline model

| Model | Random effects | Fixed effects | AIC | BIC | Log-likelihood | Test | LRT | P-value |
|---|---|---|---|---|---|---|---|---|
| M0 | (1 \| Subject) | t_c | 2610 | 2625 | -1301 | — | — | — |
| M1 | (1 + t_c \| Subject) | t_c | 2587 | 2610 | -1288 | M1 vs. M0 | 26.87 | < 0.001 |
| M2 | (1 + t_c \| Subject) | t_c * Condition | 2585 | 2614 | -1284 | M2 vs. M1 | 6.49 | 0.039 |
| M3 | (1 + t_c \| Subject) | t_c * Condition + Session | 2577 | 2611 | -1280 | M3 vs. M2 | 9.57 | 0.002 |
| M4 | (1 + t_c \| Subject) | t_c * Condition + Session * t_c | 2571 | 2609 | -1276 | M4 vs. M3 | 7.76 | 0.005 |
| M5 | (1 + t_c \| Subject) | t_c * Condition + Session * t_c + Session * Condition | 2573 | 2615 | -1276 | M5 vs. M4 | < 0.001 | 0.995 |

**Notes:** t_c denotes time centered within each working episode (t = 0) at episode onset); Condition indicates task conditions (Human-only vs. HRC); and Session denotes the two temporally ordered working periods within the experimental protocol.

Fixed effects were evaluated sequentially using LRTs. Adding Condition significantly improved model fit relative to the random-effects-only model (M2 vs. M1; LRT = 6.49, p-value = 0.039), indicating condition-dependent differences in RSME trajectories. Specifically, the time-dependent increase in RSME was larger under the HRC condition than under the Human-only condition (Δslope = 0.069). Session effects were also significant: compared to Session 1, Session 2 exhibited a lower overall RSME level (Δintercept = −9.54; M3 vs. M2; LRT = 9.57, p-value = 0.002) and a steeper increase over time (Δslope = 0.21; M4 vs. M3; LRT = 7.76, p-value = 0.005), suggesting partial adaptation in baseline cognitive strain alongside accelerated accumulation during the later working period. In addition, no significant interaction between Condition and Session was observed (M5 vs. M4; LRT < 0.001, p-value = 0.995), indicating that the effect of HRC on cognitive strain accumulation was consistent across sessions.

In practical terms, perceived cognitive strain increased approximately linearly during work, accumulated faster under the HRC condition, and showed session-related baseline and slope changes.

***4.3.2 Rest Phase: Nonlinear Recovery During Rest***

For the rest phase, exponential decay mixed-effects models were used to characterize recovery dynamics and to assess between-subject heterogeneity (random effects) and session-related differences (fixed effects). Consistent with the strong empirical support for nonlinear recovery reported in Sections 4.1–4.2, all rest-phase analyses employed the recovery form of Eq. (2). In contrast to the working phase, rest-phase trajectories provided sufficient curvature and temporal resolution to support estimation of recovery rate parameters $k$. Random-effects structures were evaluated first to establish an identifiable baseline model, after which fixed effects of Session were tested.

4.3.2.1 Physical Strain Recovery: Recovery Speed Decreases in the Second Session

For physical strain (BORG), allowing subject-specific random effects substantially improved rest-phase model fit, as shown in Table 7. Extending the random-effects structure from random effects on a and b to include a random effect on the recovery rate $k$ produced the best-supported random-effects baseline (model M4), indicating pronounced inter-individual variability not only in baseline recovery levels and amplitudes but also in recovery speed. The selected model M4 included correlated random effects on $a$, $b$, and $k$, with moderate correlation between $a$ and $b$ ( $\rho = -0.335$), $a$ positive association between $k$ and $a$ ($\rho = 0.494$), and a strong negative association between $k$ and $b$ ($\rho = -0.798$), suggesting structured coupling between recovery magnitude and recovery rate across individuals.

**Table 7.** Model selection for rest-phase physical strain (BORG) using exponential mixed-effects models, models M0–M4 evaluate random-effects structures, whereas models M5–M7 assess Session fixed effects on the selected baseline model

| Model | Random effects | Fixed effects | AIC | BIC | Log-likelihood | Test | LRT | P-value |
|---|---|---|---|---|---|---|---|---|
| M0 | None | None | 1313 | 1328 | -652 | M4 vs. M0 | 270.00 | < 0.001 |
| M1 | a | None | 1172 | 1191 | -581 | M3 vs. M1 | 78.35 | < 0.001 |
| M2 | b | None | 1147 | 1166 | -568 | M3 vs. M2 | 53.12 | < 0.001 |
| M3 | a, b | None | 1098 | 1125 | -542 | M4 vs. M3 | 48.85 | < 0.001 |
| M4 | a, b, k | None | 1055 | 1093 | -517 | — | — | — |
| M5 | a, b, k | Session → a | 1057 | 1099 | -517 | M5 vs. M4 | 0.17 | 0.683 |

| M6 | a, b, k | Session → b | 1053 | 1095 | -515 | M6 vs. M4 | 4.13 | 0.042 |
|---|---|---|---|---|---|---|---|---|
| M7 | a, b, k | Session → k | 1019 | 1062 | -499 | M7 vs. M4 | 37.52 | < 0.001 |

As shown in Figure 13, incorporating these subject-level random effects reduced systematic subject-specific residual offsets relative to the population-level model, supporting this specification as a stable baseline for subsequent fixed-effects testing.

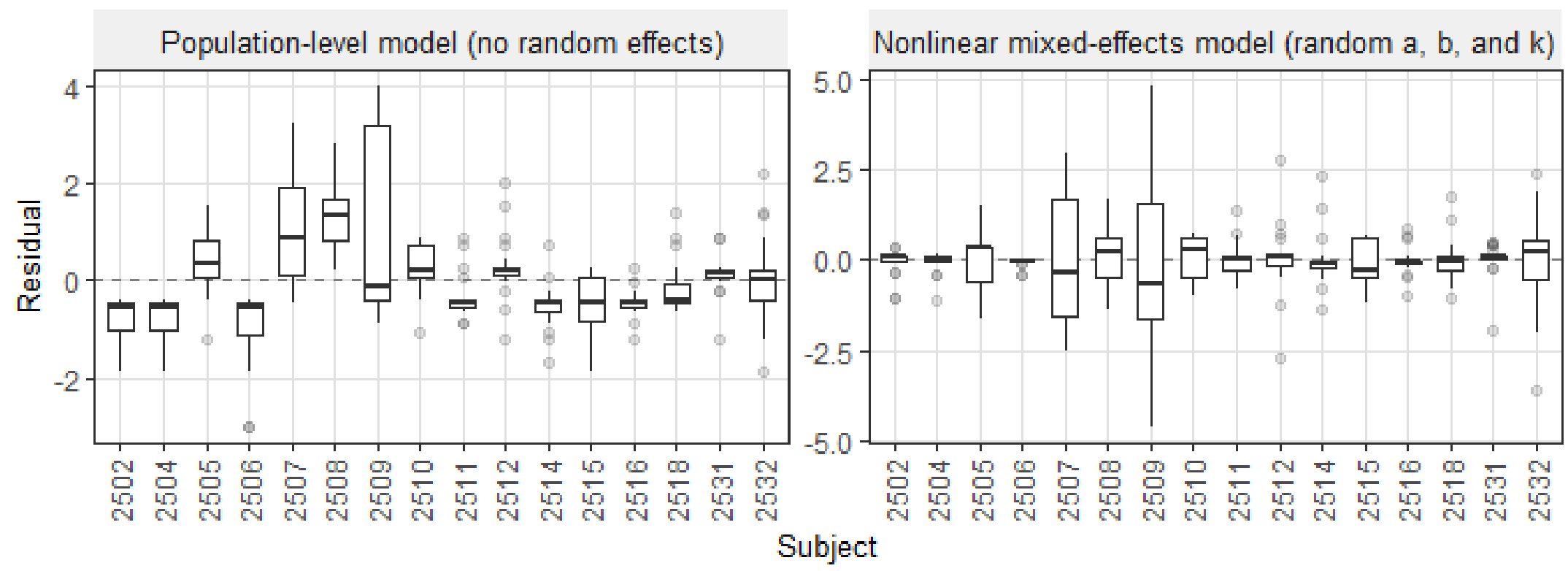

**Figure 13.** Residual distributions by subject for rest-phase physical strain (BORG), comparing the population-level exponential model and the nonlinear mixed-effects model with subject-specific random effects

Session effects were then evaluated by allowing Session to enter each recovery parameter individually while holding the random-effects structure fixed, as shown in Table 5. A strong Session effect was observed for the recovery rate parameter $k$ (M7 vs. M4; LRT = 37.52, p-value < 0.001), indicating a lower average recovery rate during the second rest period ($\Delta k = -0.09$). In contrast, adding session effects to the asymptote $a$ did not improve model fit (M5 vs. M4; LRT = 0.17, p-value = 0.683). Although allowing Session to influence the recovery amplitude $b$ yielded a nominally significant LRT (M6 vs. M4; LRT = 4.13, p-value = 0.042), the associated improvement in both AIC and log-likelihood was modest and did not translate into meaningful gains once model complexity was penalized by information criteria.

These results suggest that session-related differences in physical strain recovery are primarily captured by changes in recovery speed rather than by systematic shifts in recovery magnitude or post-recovery baseline levels.

In practical terms, perceived physical strain recovered nonlinearly during rest, but the recovery rate was lower in the second session, suggesting slower physical recovery after repeated exposure.

4.3.2.2 Cognitive Strain Recovery: Baseline Shift Rather Than Recovery-Rate Change

For cognitive strain (RSME) during the rest phase, random-effects model selection indicated substantial between-subject heterogeneity in recovery dynamics, as summarized in Table 8. Allowing subject-specific random effects on both the asymptotic parameter ($a$) and the recovery amplitude ($b$), with the recovery rate ($k$) fixed due to limited identifiability under the available temporal resolution, provided a markedly improved fit relative to models with random effects on either parameter alone. The selected baseline specification (model M3) included correlated random effects on $a$ and $b$, with a moderate positive correlation ($\rho = 0.296$), indicating that individuals with higher post-recovery asymptotic RSME levels also tended to exhibit larger recovery amplitudes.

**Table 8.** Model selection for rest-phase cognitive strain (RSME) using exponential mixed-effects models (rate parameter k fixed), models M0–M3 evaluate random-effects structures, whereas models M4–M6 assess Session fixed effects on the selected baseline model

| Model | Random effects | Fixed effects | AIC | BIC | Log-likelihood | Test | LRT | P-value |
|---|---|---|---|---|---|---|---|---|
| M0 | None | None | 2942 | 2954 | -1468 | M3 vs. M0 | 283.79 | < 0.001 |
| M1 | a | None | 2759 | 2774 | -1375 | M3 vs. M1 | 97.94 | < 0.001 |
| M2 | b | None | 2819 | 2834 | -1405 | M3 vs. M2 | 157.99 | < 0.001 |
| M3 | a, b | None | 2665 | 2688 | -1326 | — | — | — |
| M4 | a, b | Session → a | 2652 | 2679 | -1319 | M4 vs. M3 | 14.26 | < 0.001 |
| M5 | a, b | Session → b | 2664 | 2692 | -1325 | M5 vs. M3 | 2.08 | 0.15 |
| M6 | a, b | Session → a + b | 2654 | 2685 | -1319 | M6 vs. M4 | 0.34 | 0.56 |

As illustrated in Figure 14, incorporating subject-level random effects substantially reduced systematic subject-specific residual offsets compared to the population-level exponential model, yielding residuals that were more symmetrically distributed around zero across individuals. This improvement supports the selected random-effects structure as a stable and identifiable baseline for fixed-effects testing.

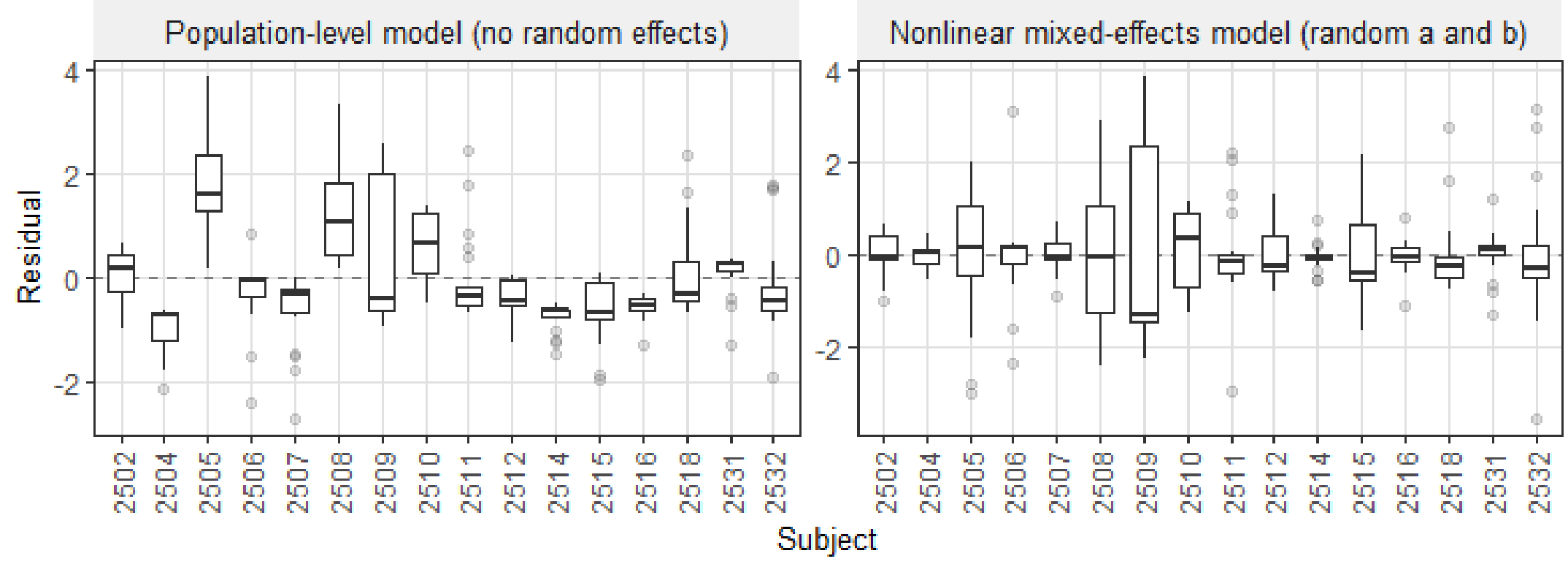

**Figure 14.** Residual distributions by subject for rest-phase cognitive strain (RSME), comparing the population-level exponential model and the nonlinear mixed-effects model with subject-specific random effects

Session effects were subsequently evaluated by allowing Session to influence the recovery parameters $a$ and $b$ while holding the random-effects structure fixed, as shown in Table 8. Session had a significant effect on the asymptotic parameter $a$ (M4 vs. M3; LRT = 14.26, p-value < 0.001). Compared to the first rest session, the second rest session was associated with a lower asymptotic RSME level ($\Delta a = -3.68$), indicating a reduced post-recovery cognitive strain state. In contrast, no significant session effect was detected for the recovery amplitude $b$ (M5 vs. M3; LRT = 2.08, p-value = 0.15) and jointly allowing Session to affect both $a$ and $b$ did not yield further improvements in model fit (M6 vs. M4; LRT = 0.34, p-value = 0.56).

In practical terms, perceived cognitive strain recovery was mainly reflected in a lower post-rest baseline in the second session rather than a clearly identifiable change in recovery speed.

## 5. DISCUSSION

### 5.1 Theoretical implications of perceived strain dynamics

The findings provide theoretical implications for understanding worker-state dynamics in HRC by positioning perceived strain as a time-dependent intermediate state between task workload and longer-term fatigue outcomes. Rather than treating worker state as a static workload score or an end-state fatigue indicator, the results show that perceived strain changes differently across work and rest phases. This supports the need to distinguish workload, perceived strain, fatigue, and recovery when modeling human state in collaborative construction tasks.

The results also suggest that perceived cognitive and physical strain should not be represented by a single generic fatigue variable. Physical strain during work was dominated by inter-individual variability, whereas cognitive strain showed clearer sensitivity to collaboration mode and repeated exposure. Recovery also showed nonlinear patterns during rest. These findings are consistent with resource depletion and recovery perspectives and provide empirical evidence that cognitive and physical worker-state dimensions may require separate phase-specific representations in HRC modeling.

### 5.2 Practical implications for HRC task allocation and scheduling

The fitted perceived strain models provide practical insights for HRC task allocation and scheduling research, particularly in contexts where productivity needs to be considered alongside physical and cognitive demands. The results suggest that future scheduling models may benefit from explicitly representing phase-dependent perceived strain accumulation and recovery, rather than relying only on static or instantaneous workload assumptions.

A first implication is the pronounced asymmetry between work and rest across both strain dimensions. During work, accumulation shows limited curvature over the observed horizon, with cognitive strain well approximated by a linear trend within the observed one-hour work period. During rest, recovery is strongly nonlinear, with rapid early relief followed by diminishing returns. Consequently, rest may not be adequately represented as a simple reset: timing and duration may need to be considered explicitly, since short breaks may yield large benefits while longer breaks add progressively less. This supports the potential value of nonlinear recovery constraints in scheduling models rather than fixed rest quotas.

Second, physical strain accumulation during work is dominated by inter-individual variability. The working-phase BORG models indicate substantial subject heterogeneity and no supported fixed effects of condition or session. For scheduling, this suggests that physical task allocation may need to account for individual capacity profiles rather than relying on global task-level rules. Uniform physical workload limits or condition-based adjustments may be insufficient in some contexts; subject-specific constraints or adaptive assignment policies could provide more flexible alternatives.

Cognitive strain, in contrast, is sensitive to collaboration mode and temporal exposure. Working-phase RSME shows higher accumulation rates under HRC and session-dependent changes: Session 2 starts lower but increases faster. This indicates that adaptation and fatigue can coexist—operators may begin later sessions with reduced perceived cognitive strain yet accumulate more quickly. Together with the nonlinear recovery dynamics observed during rest, these results suggest the importance of modeling both rapidly recoverable and slowly accumulating cognitive fatigue components. Scheduling models may therefore benefit from considering exposure duration in cognitively demanding HRC tasks rather than relying on baseline workload alone, for example by adjusting task length or inserting cognitive rest based on elapsed collaboration time.

Importantly, these models may be interpreted as candidate model structures for future optimization-based task allocation and scheduling. As summarized in Table 9, the finalized phase- and scale-specific functional forms and parameters may serve as accumulation/recovery cost functions or constraints on allowable work duration and required recovery. Even when parameters do not transfer, the identified model structures, including near-linear accumulation during work and nonlinear decay during rest, may provide useful starting templates that can be recalibrated with new task-specific data. These results provide a basis for future integration of perceived strain dynamics into HRC scheduling, although additional validation would be required for new tasks, worker populations, and operational settings.

**Table 9.** Summary of physical (BORG) and cognitive (RSME) strain accumulation and recovery models across phases and scales

| Phase | Scale | Model form | Random effects | Fixed-effects parameters | Supported fixed effects | Interpretation |
|---|---|---|---|---|---|---|
| Working | BORG | Exponential accumulation Eq. (1) | a, b | a = 8.247<br>b = 3.008<br>k = 0.030 min$^{-1}$ (fixed) | None | Physical strain accumulation was dominated by inter-individual variability; no systematic condition- or session-level effects were supported. |
| Working | RSME | Linear Eq. (3) | β0, β1 | β0 = 48.817<br>β1 = 0.090 min$^{-1}$ | Condition → slope<br>Session → intercept & slope | HRC showed a positive slope difference (Δslope_HRC = +0.069 min$^{-1}$), although the 95% CI included zero. Session 2 exhibited a lower baseline (Δintercept_S2 = −9.54) but a faster accumulation rate (Δslope_S2 = +0.21 min$^{-1}$). |
| | | Exponential accumulation Eq. (1) | — | a = 39.837<br>b = 18.613<br>k = 0.031 min$^{-1}$ (ns) | — | Optional nominal representation for long-horizon planning when plateau behavior needs to be represented; not used for inferential purposes due to weak identifiability of k. |
| Rest | BORG | Exponential decay Eq. (2) | a, b, k | a = 6.593<br>b = 4.223<br>k = 0.331 min$^{-1}$ | Session → k | Physical recovery rate decreased in Session 2 (Δk_S2 = −0.09 min$^{-1}$). |
| Rest | RSME | Exponential decay Eq. (2) | a, b | a = 11.869<br>b = 47.740<br>k = 0.237 min$^{-1}$ (fixed) | Session → a | Cognitive recovery was primarily expressed as a shift in post-recovery baseline (Δa_S2 = −3.68), rather than changes in recovery speed. |

**Notes**: BORG ranges from 6–20 and RSME from 0–150, with higher values indicating greater physical exertion and mental strain, perspectively. The fixed $k$ in the final Rest RSME model was based on early rest-phase behavior and differs from the full population-level estimate in Table 4.

### 5.3 Methodological implications for model-identifiable HRC experiments

These findings also provide methodological implications for designing HRC experiments that support dynamic human-state modeling. Many existing HRC experiments may be less suited for dynamic modeling because they rely on single-task trials, short observation windows, or static workload indicators [20–22]. Such designs are effective for comparing instantaneous performance or perceived workload but may provide limited information about accumulation and recovery processes that unfold over time.

The protocol developed in this study responds to this limitation by aligning experimental structure with modeling requirements. Repeated work–rest cycles can create sufficient temporal contrast to distinguish accumulation from recovery, while the alternation between cognitively and physically demanding tasks partially reflects the complexity of real-world prefabrication activities. Importantly, the protocol is not designed merely to collect human-factor measurements, but to support identifiable dynamic models that may later inform task allocation and scheduling formulations after validation. This modeling-oriented perspective differentiates the protocol from many prior HRC studies in prefabricated construction, where data collection and analysis are often decoupled from downstream decision-making needs.

Figure 15 illustrates how this protocol can be abstracted into an adaptable, human-centric experimental design framework spanning three interrelated dimensions. The task dimension emphasizes abstraction and decomposition of real-world prefabrication workflows, followed by scaling and simplification to achieve experimental tractability while preserving essential physical and cognitive demands. The human dimension focuses on explicitly specifying the types of abilities engaged by the task, such as sustained physical effort versus attentional and coordination demands, and selecting measurement variables that are sensitive to temporal dynamics. The robot dimension captures how robot roles, interaction patterns, and control strategies shape human effort, allowing systematic comparison between human-only and HRC conditions. Structuring experiments around these dimensions and their associated design logic, researchers may improve the suitability of collected data for temporal analysis and model-based interpretation.

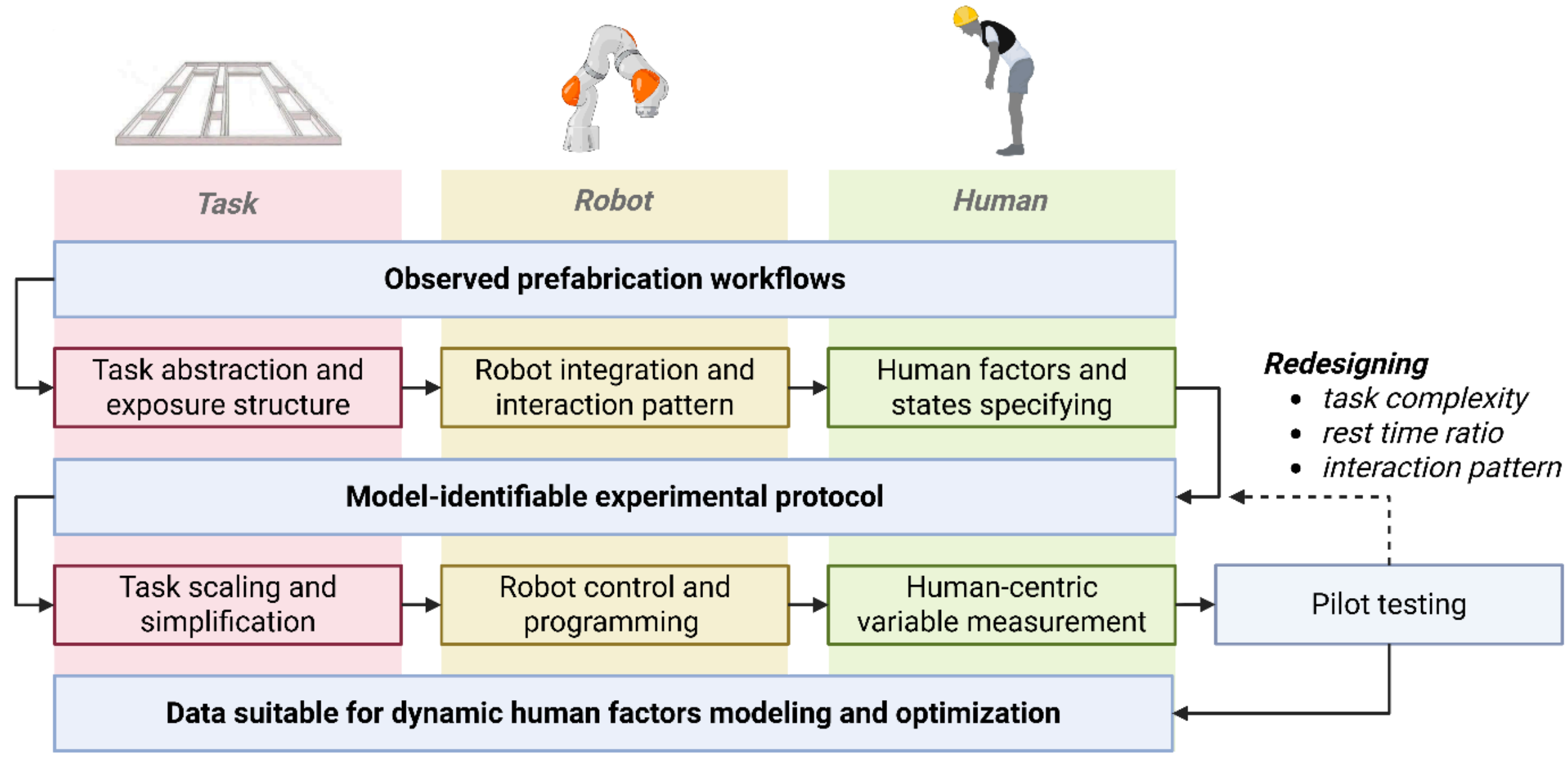

**Figure 15.** Human-centric experimental design framework for model-identifiable HRC studies in prefabricated construction

The framework is intended to be extensible within related controlled HRC experimental settings. While the present study employed two work–rest sessions with fixed collaboration modes, the same structure can accommodate longer work durations, additional sessions, or adaptive HRC strategies that evolve in response to measured strain. Such extensions may be possible through adjustments to task duration, repetition, or collaboration details. In this sense, the framework provides a methodological scaffold rather than a task-specific recipe. By foregrounding human-state variables and their temporal evolution, it offers a methodological scaffold for future HRC studies to generate temporally structured data that can support dynamic modeling and inform later optimization and human-aware system design after context-specific validation.

**5.4 Limitations and future work**

Several limitations should be acknowledged. First, the controlled laboratory setup may not fully represent the environmental variability, interruptions, and task diversity present in real prefabrication sites. Second, the participant sample consisted primarily of current students and recent graduates rather than professional construction workers. Therefore, the estimated perceived-strain trajectories and model parameters should be interpreted as context-specific and may not directly generalize to experienced prefabrication workers. In addition, the limited sample size may affect between-subject generalizability and the stability of more complex nonlinear mixed-effects specifications. Third, although validated self-report scales were used, perceived strain ratings may still be affected by subjective reporting bias and do not fully capture physiological or behavioral responses associated with sustained work and recovery. Therefore, the identified model structures should be further validated and recalibrated with larger and more diverse samples, professional workers and field or factory data.

Several directions for future work emerge from this study. The experimental protocol can be extended to longer work durations, additional sessions, or more diverse collaboration strategies to examine longer-term adaptation, fatigue outcomes, and performance decline. Future studies may also incorporate physiological or behavioral indicators as complementary measures to examine how perceived strain relates to objective worker-state responses. Finally, evaluating whether the identified perceived-strain model structures can be embedded into real-time optimization or control frameworks represents an important direction for future work toward balancing productivity, safety, and human sustainability in prefabricated construction and related domains.

## 6. CONCLUSIONS

This study investigated the temporal dynamics of human state variables in HRC within a prefabrication context, with a focus on how perceived cognitive and physical strain accumulates during work and recovers during rest. By combining repeated work–rest experimental protocols with statistical modeling, the results suggest that perceived strain exhibits clear phase-dependent behaviors, with distinct accumulation and recovery dynamics that vary across physical and cognitive dimensions, collaboration modes, and repeated exposure. These findings indicate the value of considering time, exposure, and recovery when evaluating perceived strain in HRC systems.

This work makes three primary contributions. First, it provides empirically grounded models of perceived physical and cognitive strain accumulation and recovery across working and resting phases, supporting a time-dependent representation of worker state within the studied task context. Second, it shows how these fitted models may inform future HRC task allocation and scheduling research and human-aware decision models by representing dynamic perceived strain patterns rather than relying on static thresholds. Third, it introduces a human-centric experimental protocol that aligns task design, collaboration structure, and measurement selection with the requirements of dynamic modeling, offering a framework that may guide future HRC studies.

## DATA STATEMENT

Data will be made available on request.

## REFERENCES

[1] J. Du, J. Zhang, D. Castro-Lacouture, Y. Hu, Lean manufacturing applications in prefabricated construction projects, Autom. Constr. 150 (2023) 104790. https://doi.org/10.1016/J.AUTCON.2023.104790.

[2] V. Tavares, N. Soares, N. Raposo, P. Marques, F. Freire, Prefabricated versus conventional construction: Comparing life-cycle impacts of alternative structural materials, Journal of Building Engineering 41 (2021) 102705. https://doi.org/10.1016/J.JOBE.2021.102705.

[3] Y. Fu, J. Chen, W. Lu, Human-robot collaboration for modular construction manufacturing: Review of academic research, Autom. Constr. 158 (2024) 105196. https://doi.org/10.1016/J.AUTCON.2023.105196.

[4] A. Zhu, P. Pauwels, B. De Vries, Smart component-oriented method of construction robot coordination for prefabricated housing, Autom. Constr. 129 (2021) 103778. https://doi.org/10.1016/J.AUTCON.2021.103778.

[5] G. Kokotinis, G. Michalos, Z. Arkouli, S. Makris, On the quantification of human-robot collaboration quality, Int. J. Comput. Integr. Manuf. 36 (2023) 1431–1448. https://doi.org/10.1080/0951192X.2023.2189304.
[6] B.M. Tehrani, A. Alwisy, Optimizing task allocation in human-in-the-lead construction robotics: A framework for wood assembly–based robotics in panelized construction, Journal of Computing in Civil Engineering 39 (2025) 04025097. https://doi.org/10.1061/JCCEE5.CPENG-6570.
[7] T. Yu, J. Huang, Q. Chang, Optimizing task scheduling in human-robot collaboration with deep multi-agent reinforcement learning, J. Manuf. Syst. 60 (2021) 487–499. https://doi.org/10.1016/J.JMSY.2021.07.015.
[8] M.L. Lee, S. Behdad, X. Liang, M. Zheng, Task allocation and planning for product disassembly with human–robot collaboration, Robot. Comput. Integr. Manuf. 76 (2022) 102306. https://doi.org/10.1016/J.RCIM.2021.102306.
[9] R. Zhao, S. Tao, P. Li, Human-centric assembly planning framework for human–robot collaborative systems with efficient reinforcement self-learning multi-objective evolutionary optimizer, Advanced Engineering Informatics 67 (2025) 103508. https://doi.org/10.1016/J.AEI.2025.103508.
[10] S. Pütz, A. Mertens, L.L. Chuang, V. Nitsch, Physiological predictors of operator performance: The role of mental effort and its link to task performance, Hum. Factors 67 (2025) 595–615. https://doi.org/10.1177/00187208241296830.
[11] Y. Liu, M. Habibnezhad, H. Jebelli, Brainwave-driven human-robot collaboration in construction, Autom. Constr. 124 (2021) 103556. https://doi.org/10.1016/J.AUTCON.2021.103556.
[12] Y. Wang, C. Mao, T. Wang, X. Tao, Enhancing construction safety through hazard-aware human–robot task allocation, Journal of Management in Engineering 42 (2026) 04026002. https://doi.org/10.1061/JMENEA.MEENG-7134.
[13] A. Adel, D. Ruan, W. McGee, S. Mozaffari, Feedback-driven adaptive multi-robot timber construction, Autom. Constr. 164 (2024) 105444. https://doi.org/10.1016/J.AUTCON.2024.105444.
[14] S. Chand, Y. Lu, Dual task scheduling strategy for personalized multi-objective optimization of cycle time and fatigue in human-robot collaboration, Manuf. Lett. 35 (2023) 88–95. https://doi.org/10.1016/J.MFGLET.2023.08.064.
[15] B. Yao, X. Li, Z. Ji, K. Xiao, W. Xu, Task reallocation of human-robot collaborative production workshop based on a dynamic human fatigue model, Comput. Ind. Eng. 189 (2024) 109855. https://doi.org/10.1016/J.CIE.2023.109855.
[16] C. Petzoldt, M. Harms, M. Freitag, Review of task allocation for human-robot collaboration in assembly, Int. J. Comput. Integr. Manuf. 36 (2023) 1675–1715. https://doi.org/10.1080/0951192X.2023.2204467.
[17] F.A. Haji, D. Rojas, R. Childs, S. de Ribaupierre, A. Dubrowski, Measuring cognitive load: Performance, mental effort and simulation task complexity, Med. Educ. 49 (2015) 815–827. https://doi.org/10.1111/MEDU.12773.
[18] M.S. Young, K.A. Brookhuis, C.D. Wickens, P.A. Hancock, State of science: mental workload in ergonomics, Ergonomics 58 (2015) 1–17. https://doi.org/10.1080/00140139.2014.956151.
[19] J. Flanagan, D. Nathan-Roberts, Theories of vigilance and the prospect of cognitive restoration, Proceedings of the Human Factors and Ergonomics Society 63 (2019) 1639–1643. https://doi.org/10.1177/1071181319631506.
[20] Y. Wang, X. Hou, B. Xiao, S.T. Mueller, Understanding cognitive impacts of robots in worker-robot collaboration in modular construction, in: 42nd International Symposium on

Automation and Robotics in Construction, Montreal, Canada, 2025: pp. 837–844. https://doi.org/10.22260/ISARC2025/0109.
[21] M. Capponi, R. Gervasi, L. Mastrogiacomo, F. Franceschini, Assembly complexity and physiological response in human-robot collaboration: Insights from a preliminary experimental analysis, Robot. Comput. Integr. Manuf. 89 (2024) 102789. https://doi.org/10.1016/J.RCIM.2024.102789.
[22] B. Sawicki, P. Düking, G. Placzek, L. Masur, R. Dörrie, P. Schwerdtner, H. Kloft, Human–robot collaboration in digital fabrication with concrete: quantifying productivity and psychophysiological strain of human workers, Construction Robotics 10 (2026) 4. https://doi.org/10.1007/S41693-025-00173-X.
[23] C.Z. Li, F. Xue, X. Li, J. Hong, G.Q. Shen, An Internet of Things-enabled BIM platform for on-site assembly services in prefabricated construction, Autom. Constr. 89 (2018) 146–161. https://doi.org/10.1016/J.AUTCON.2018.01.001.
[24] C. Liu, J. Wu, X. Jiang, Y. Gu, L. Xie, Z. Huang, Automatic assembly of prefabricated components based on vision-guided robot, Autom. Constr. 162 (2024) 105385. https://doi.org/10.1016/J.AUTCON.2024.105385.
[25] C.-H. Yang, L.-T. Tsai, Y. Chen, S.-C. Kang, Assembly sequence planning method for robotic timber wall prefabrication, Autom. Constr. 181 (2026) 106669. https://doi.org/10.1016/J.AUTCON.2025.106669.
[26] M. Marchesi, D.T. Matt, Design for mass customization: Rethinking prefabricated housing using axiomatic design, Journal of Architectural Engineering 23 (2017) 05017004. https://doi.org/10.1061/(ASCE)AE.1943-5568.0000260.
[27] X. Yang, F. Amtsberg, M. Sedlmair, A. Menges, Challenges and potential for human–robot collaboration in timber prefabrication, Autom. Constr. 160 (2024) 105333. https://doi.org/10.1016/J.AUTCON.2024.105333.
[28] Y. Wang, X. Hou, T. Chen, B. Xiao, Recent advancements of applied robotics in construction project management: A life cycle perspective, ASCE Open: Multidisciplinary Journal of Civil Engineering 3 (2025) 03125002. https://doi.org/10.1061/AOMJAH.AOENG-0095.
[29] L.-T. Tsai, C.-H. Yang, Y. Chen, Y.H. Chui, Development of a human–robot collaborative process design framework and an associated planning tool for robotic building prefabrication, Construction Robotics 10 (2026) 8. https://doi.org/10.1007/S41693-026-00181-5.
[30] Y. Fu, W. Lu, Design considerations of ergonomic human-robot collaboration (EHRC) for modular construction manufacturing, in: 42nd International Symposium on Automation and Robotics in Construction (ISARC 2025), IAARC Publications, Montreal, Canada, 2025: pp. 917–924. https://doi.org/10.22260/ISARC2025/0119.
[31] S. Li, P. Zheng, S. Liu, Z. Wang, X.V. Wang, L. Zheng, L. Wang, Proactive human–robot collaboration: Mutual-cognitive, predictable, and self-organising perspectives, Robot. Comput. Integr. Manuf. 81 (2023) 102510. https://doi.org/10.1016/J.RCIM.2022.102510.
[32] F. Ranz, V. Hummel, W. Sihn, Capability-based task allocation in human-robot collaboration, Procedia Manuf. 9 (2017) 182–189. https://doi.org/10.1016/J.PROMFG.2017.04.011.
[33] P. Tsarouchi, A.S. Matthaiakis, S. Makris, G. Chryssolouris, On a human-robot collaboration in an assembly cell, Int. J. Comput. Integr. Manuf. 30 (2017) 580–589. https://doi.org/10.1080/0951192X.2016.1187297.
[34] M.Y. Jaber, Z.S. Givi, W.P. Neumann, Incorporating human fatigue and recovery into the learning–forgetting process, Appl. Math. Model. 37 (2013) 7287–7299. https://doi.org/10.1016/J.APM.2013.02.028.

[35] A. Baratta, A. Cimino, F. Longo, G. Mirabelli, L. Nicoletti, Task allocation in human-robot collaboration: A simulation-based approach to optimize operator's productivity and ergonomics, Procedia Comput. Sci. 232 (2024) 688–697. https://doi.org/10.1016/J.PROCS.2024.01.068.
[36] C.D. Wickens, Multiple resources and mental workload, Hum. Factors 50 (2008) 449–455. https://doi.org/10.1518/001872008X288394.
[37] C.D. Wickens, Multiple resources and performance prediction, Theor. Issues Ergon. Sci. 3 (2002) 159–177. https://doi.org/10.1080/14639220210123806.
[38] B.M. Gabriel, J.R. Zierath, The limits of exercise physiology: From performance to health, Cell Metab. 25 (2017) 1000–1011. https://doi.org/10.1016/J.CMET.2017.04.018.
[39] G. Robert, J. Hockey, Compensatory control in the regulation of human performance under stress and high workload: A cognitive-energetical framework, Biol. Psychol. 45 (1997) 73–93. https://doi.org/10.1016/S0301-0511(96)05223-4.
[40] D. van der Linden, M. Frese, T.F. Meijman, Mental fatigue and the control of cognitive processes: effects on perseveration and planning, Acta Psychol. (Amst). 113 (2003) 45–65. https://doi.org/10.1016/S0001-6918(02)00150-6.
[41] O. Chen, J.C. Castro-Alonso, F. Paas, J. Sweller, Extending cognitive load theory to incorporate working memory resource depletion: Evidence from the spacing effect, Educational Psychology Review 30 (2018) 483–501. https://doi.org/10.1007/S10648-017-9426-2.
[42] W. Macdonald, The impact of job demands and workload on stress and fatigue, Aust. Psychol. 38 (2003) 102–117. https://doi.org/10.1080/00050060310001707107.
[43] Y. Wang, Y. Huang, B. Gu, S. Cao, D. Fang, Identifying mental fatigue of construction workers using EEG and deep learning, Autom. Constr. 151 (2023) 104887. https://doi.org/10.1016/J.AUTCON.2023.104887.
[44] A. Aryal, A. Ghahramani, B. Becerik-Gerber, Monitoring fatigue in construction workers using physiological measurements, Autom. Constr. 82 (2017) 154–165. https://doi.org/10.1016/J.AUTCON.2017.03.003.
[45] J.F. Hopstaken, D. van der Linden, A.B. Bakker, M.A.J. Kompier, Y.K. Leung, Shifts in attention during mental fatigue: Evidence from subjective, behavioral, physiological, and eye-tracking data, J. Exp. Psychol. Hum. Percept. Perform. 42 (2016) 878–889. https://doi.org/10.1037/XHP0000189.
[46] R.L. Charles, J. Nixon, Measuring mental workload using physiological measures: A systematic review, Appl. Ergon. 74 (2019) 221–232. https://doi.org/10.1016/J.APERGO.2018.08.028.
[47] G.A. Borg, Psychophysical bases of perceived exertion, Med. Sci. Sports Exerc. 14 (1982) 377–381. https://doi.org/10.1249/00005768-198205000-00012.
[48] F.R.H. Zijlstra, Efficiency in work behaviour: A design approach for modern tools, Delft University Press, Delft, 1993.
[49] S.G. Hart, L.E. Staveland, Development of NASA-TLX (Task Load Index): Results of empirical and theoretical research, Advances in Psychology 52 (1988) 139–183. https://doi.org/10.1016/S0166-4115(08)62386-9.
[50] F.G.W.C. Paas, Training strategies for attaining transfer of problem-solving skill in statistics: A cognitive-load approach, J. Educ. Psychol. 84 (1992) 429–434. https://doi.org/10.1037/0022-0663.84.4.429.
[51] A. Widyanti, A. Johnson, D. de Waard, Adaptation of the Rating Scale Mental Effort (RSME) for use in Indonesia, Int. J. Ind. Ergon. 43 (2013) 70–76. https://doi.org/10.1016/J.ERGON.2012.11.003.

[52] J. Leppink, P. Pérez-Fuster, Mental effort, workload, time on task, and certainty: Beyond linear models, Educational Psychology Review 31 (2019) 421–438. https://doi.org/10.1007/S10648-018-09460-2.
[53] A. Szulewski, A. Gegenfurtner, D.W. Howes, M.L.A. Sivilotti, J.J.G. van Merriënboer, Measuring physician cognitive load: validity evidence for a physiologic and a psychometric tool, Advances in Health Sciences Education 22 (2017) 951–968. https://doi.org/10.1007/S10459-016-9725-2.
[54] Y. Wang, B. Xiao, S.T. Mueller, W. Yi, Task–Taxon–Task framework for modeling and predicting the cognitive impact of collaborative robots on worker performance in modular construction, Journal of Construction Engineering and Management 152 (2026) 04026033. https://doi.org/10.1061/JCEMD4.COENG-17137.
[55] S. Papegaaij, T. Hortobágyi, B. Godde, W.A. Kaan, P. Erhard, C. Voelcker-Rehage, Neural correlates of motor-cognitive dual-tasking in young and old adults, PLoS One 12 (2017) e0189025. https://doi.org/10.1371/JOURNAL.PONE.0189025.
[56] A.M. Treisman, G. Gelade, A feature-integration theory of attention, Cogn. Psychol. 12 (1980) 97–136. https://doi.org/10.1016/0010-0285(80)90005-5.
[57] S.T. Mueller, B.J. Piper, The Psychology Experiment Building Language (PEBL) and PEBL Test Battery, J. Neurosci. Methods 222 (2014) 250–259. https://doi.org/10.1016/j.jneumeth.2013.10.024.
[58] R Core Team, R: A language and environment for statistical computing, R Foundation for Statistical Computing, Vienna, Austria, 2024. https://www.R-project.org/.